\documentclass{egpubl}
\usepackage{eg2023}
 
\ConferencePaper        %

\usepackage[T1]{fontenc}
\usepackage{dfadobe}  

\biberVersion
\BibtexOrBiblatex
\usepackage[backend=biber,bibstyle=EG,citestyle=alphabetic,backref=true]{biblatex} 
\electronicVersion
\PrintedOrElectronic

\ifpdf \usepackage[pdftex]{graphicx} \pdfcompresslevel=9
\else \usepackage[dvips]{graphicx} \fi

\usepackage{egweblnk} 
\usepackage{tikz}
\usepackage{comment}
\usepackage{amsmath,amssymb} %
\usepackage{color}
\usepackage[accsupp]{axessibility}  %

\usepackage{booktabs}
\usepackage{tabularx}
\usepackage{multirow}
\usepackage{xspace}
\usepackage{subfigure}
\usepackage{stackengine}

\newcommand{\hadar}[1]{{\textcolor{blue}{[Hadar: #1]}}}

\long\def\ignorethis#1{}

\newcommand{\datasetname}{\emph{Faces Through Time}\xspace}
\newcommand{\datasetshortname}{FTT\xspace}
\newcommand{\methodshort}{TMT\xspace}

\DeclareMathOperator*{\argmin}{arg\,min}

\newbox\jsavebox
\newcommand{\jsubfig}[2]{%
	\sbox\jsavebox{#1}%
	\parbox[t]{\wd\jsavebox}{\centering\usebox\jsavebox\\#2}%
	}

\newcolumntype{F}{>{\centering\arraybackslash}X}

\usepackage[capitalize]{cleveref}
\crefname{section}{Sec.}{Secs.}
\Crefname{section}{Section}{Sections}
\Crefname{table}{Table}{Tables}
\crefname{table}{Tab.}{Tabs.}

\title[Faces Through Time]{What's in a Decade? Transforming Faces Through Time} 

\author[E. M. Chen et al.,]
{\parbox{\textwidth}{\centering Eric Ming Chen$^{1}$,\: Jin Sun$^{2}$,\: Apoorv Khandelwal$^{3}$,\: Dani Lischinski$^{4}$,\: Noah Snavely$^{1}$, \: Hadar Averbuch-Elor$^{5}$ 
        }
\\
{\parbox{\textwidth}{\centering $^1$Cornell University, Cornell Tech\:
        $^2$University of Georgia \:
        $^3$Brown University\:
        $^4$The Hebrew University of Jerusalem\:
        $^5$Tel Aviv University\:
       }
}
}

\begin{document}

\teaser{
\centering
\begin{minipage}{0.49\linewidth}
\jsubfig{\includegraphics[width=\textwidth]{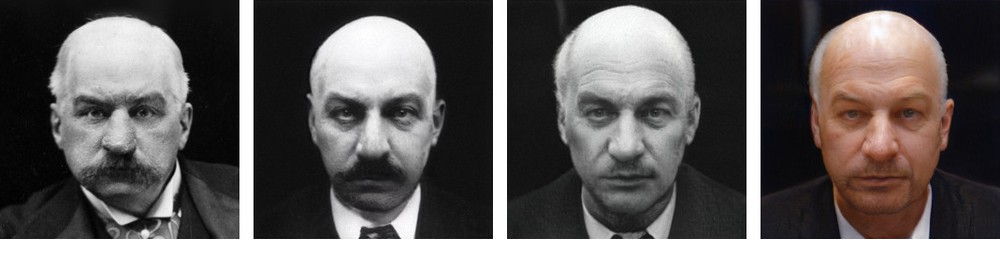}}{}
\jsubfig{\includegraphics[width=\textwidth]{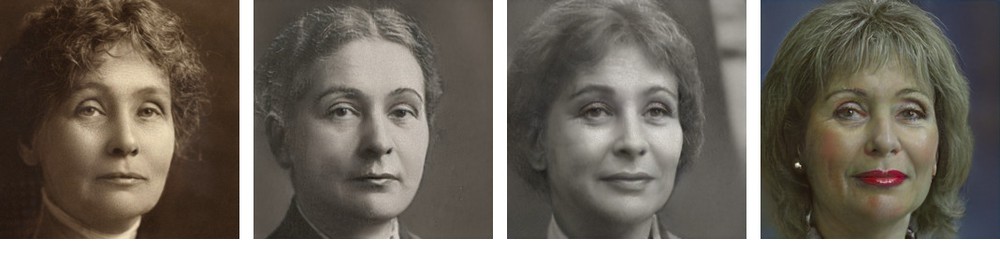}}{}
\jsubfig{\includegraphics[width=\textwidth]{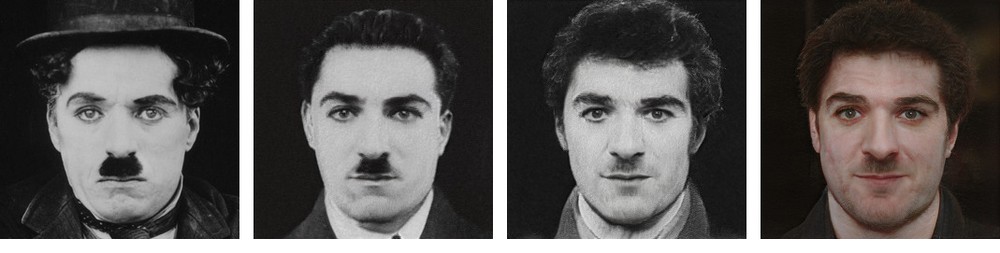}}{}
\begin{tabularx}{\linewidth}{FFFF}
Input (1910s)  & 1930s & 1970s & 2000s
\end{tabularx}
\end{minipage}
\hfill
\begin{minipage}{0.49\linewidth}
\jsubfig{\includegraphics[width=\textwidth]{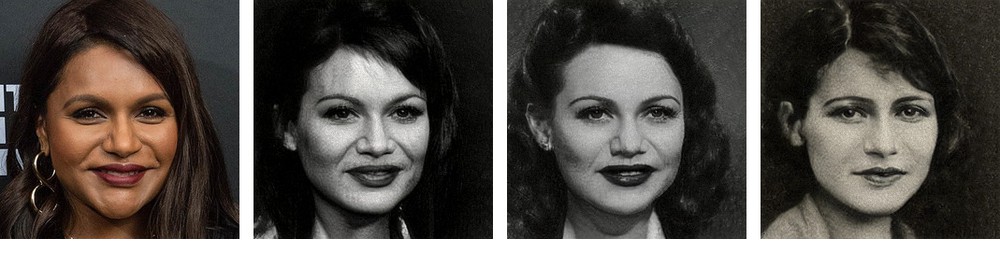}}{}
\jsubfig{\includegraphics[width=\textwidth]{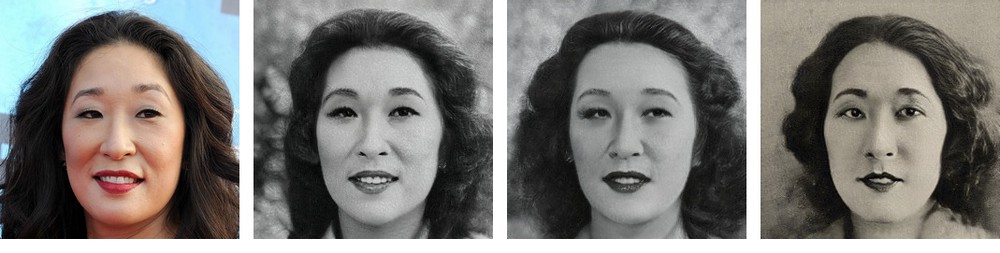}}{}
\jsubfig{\includegraphics[width=\textwidth]{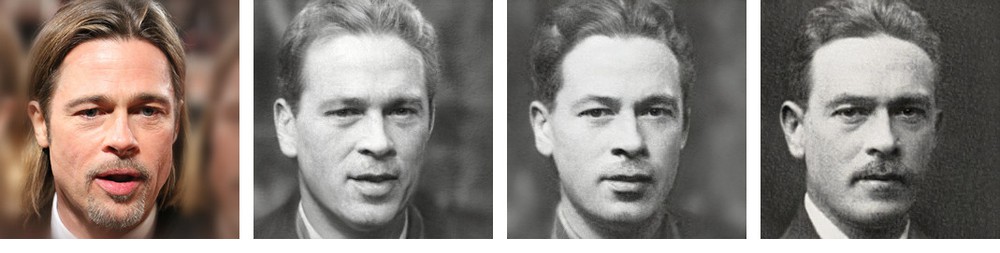}}{}
\begin{tabularx}{\linewidth}{FFFF}
Input (2010s)  & 1960s & 1940s & 1920s
\end{tabularx}
\end{minipage}
\caption{
Given an input portrait image, %
our method generates plausible renditions of what that portrait might have looked like had it been taken in different decades spanning over a century. Our framework captures characteristic styles across decades while maintaining the person's identity. From top left to bottom right: J.P.~Morgan, Mindy Kaling, Emmeline Pankhurst, Sandra Oh, Charlie Chaplin, Brad Pitt.
}
  \label{fig:intro_examples}
}
\maketitle
\begin{abstract}
How can one visually characterize photographs of people over time? In this work, we describe the \datasetname{} dataset, which contains over a thousand portrait images per decade from the 1880s to the present day. Using our new dataset, we devise a framework for resynthesizing portrait images across time, imagining how a portrait taken during a particular decade might have looked like had it been taken in other decades. Our framework optimizes a family of per-decade generators that reveal subtle changes that differentiate decades---such as different hairstyles or makeup---while maintaining the identity of the input portrait. Experiments show that our method can more effectively resynthesizing portraits across time compared to state-of-the-art image-to-image translation methods, as well as attribute-based and language-guided portrait editing models. Our code and data will be available at  \href{https://facesthroughtime.github.io}{\url{https://facesthroughtime.github.io}}. %

\ccsdesc[500]{Computing methodologies~Image manipulation}
\ccsdesc[100]{Computing methodologies~Computer graphics}
\ccsdesc[100]{Computing methodologies~Computer vision}

\printccsdesc   
\end{abstract}
\section{Introduction}
What would photographs of ourselves look like if we were born fifty, sixty, or a hundred years ago? 
What would Charlie Chaplin look like if he were active in the 2020s instead of the 1920s?
Such is the conceit of popular diversions like \href{https://en.wikipedia.org/wiki/Old-time_photography}{old-time photography}, where we imagine ourselves as we might have looked in an anachronistic time period like the Roaring Twenties.
However, while many methods for editing portraits have been devised recently on the basis of powerful generative techniques like StyleGAN~\cite{karras2020analyzing,Karras2020ada,shen2020interfacegan,10.1145/3447648,Wu2021StyleSpaceAD,alaluf2021matter,or2020lifespan}, little attention has been paid to the problem of automatically translating portrait imagery in time, while preserving other aspects of the person portrayed. This paper addresses this problem, producing results like those shown in \Cref{fig:intro_examples}.

To simulate such a ``time travel'' effect, we must be able to model and apply the characteristic 
features of a certain era. 
Such features may include stylistic trends in clothing, hair, and makeup, as well as the imaging characteristics of the cameras and film of the day. 
In this way, translating imagery across time differs from standard portrait editing effects that typically manipulate well-defined semantic attributes (\emph{e.g.}, adding or removing a smile or modifying the subject's age). Further, while large amounts of data capturing these attributes are readily available via datasets like Flickr-Faces-HQ (FFHQ)~\cite{karras2019style}, diverse and high-quality imagery spanning the history of photography is comparatively scarce.
\begin{figure*}[t]
\vspace{-4pt}
 \centering %
 
\rotatebox{90}{\begin{tabularx}{188px}{FFFFF}1990s & 1970s & 1940s & 1910s & 1880s\end{tabularx}}
\jsubfig{\includegraphics[width=0.9\textwidth]{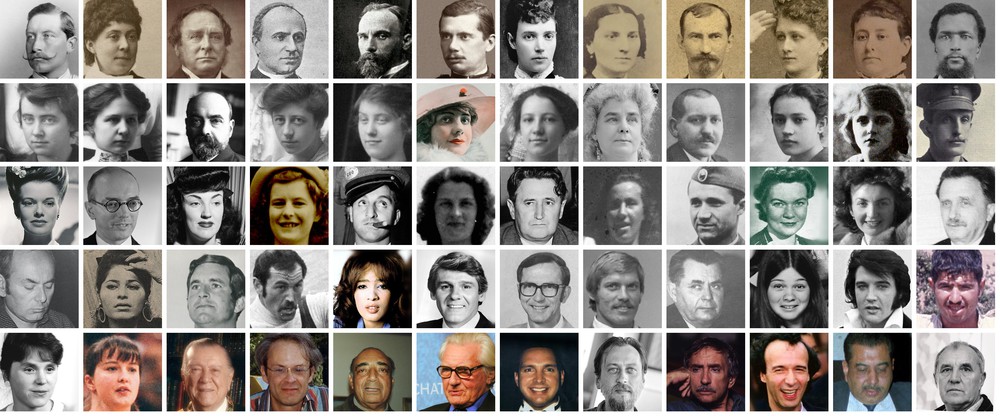}}{}
\caption{Random samples from five decades in the \datasetname{} Dataset. 
 }
  \label{fig:samples}
\end{figure*}

In this work, we take a step towards transforming images of people across time, focusing on portraits. We introduce \datasetname{} (\datasetshortname{}), a dataset containing thousands of images spanning fourteen decades from the 1880s to the present day.  \datasetname{} is derived from the massive public catalog of freely-licensed images and annotations available through the \href{https://commons.wikimedia.org}{Wikimedia Commons} project. The extensive biographic depth available on Wikimedia Commons, as well as its organization into time-based categories, enables associating images with accurate time labels. %
In comparison to previous time-stamped portrait datasets, 
\datasetshortname is sourced from a wider assortment of images capturing notable identities varying in age, nationality, pose, etc. In contrast, a well-known prior dataset called The Yearbook Dataset~\cite{ginosar2015century} only contains US yearbook photos. In our work, we demonstrate \datasetshortname's applicability for synthesis tasks. More broadly, the dataset can allow for exploration of a variety of analysis tasks, such as estimating time from images, understanding photographic styles across time, and discovering fashion trends (see Section \ref{sec:related}). Figure \ref{fig:samples} shows random samples from five different decades from \datasetshortname. 

To transform portraits across time, we build on the success of Generative Adversarial Networks (GANs) for synthesizing high-quality
facial images. 
In particular, we finetune the popular StyleGAN2~\cite{karras2020analyzing,Karras2020ada} generator network (trained on FFHQ) on our dataset. 
However, rather than modeling the entire image distribution of our dataset using a single StyleGAN2 model, we train a separate model for each decade. 
We introduce a method to align and map a person's image across the latent generative spaces of the fourteen different decades (see Figure \ref{fig:model_family}).  
Furthermore, we discover a remarkable linearity in each model's generator weights, allowing us to fine-tune images with
vector arithmetic on the model weights. 
This sets our approach apart from the many prior works that search for editing directions within a single StyleGAN2 model, \cite{shen2020interfacegan,shen2021closed,Wu2021StyleSpaceAD,10.1145/3447648,alaluf2021matter,Patashnik_2021_ICCV}.  We find that by using multiple StyleGAN2 models in this way, our method is more expressive than these existing approaches. In addition, our classes are separated in a useful way for style transfer.

We demonstrate results for a variety of individuals from different backgrounds and captured during different decades. We show that prior methods struggle on our problem setting, even when trained directly on our dataset. We also perform a quantitative evaluation, demonstrating that our transformed images are of high quality and resemble the target decades, while preserving the input identity. Our approach enables an \emph{Analysis by Synthesis} scheme that can reveal and visualize differences between portraits across time, enabling a study of fashion trends or other visual cultural elements (related to, for instance, a \href{https://www.nytimes.com/2021/05/14/us/george-washington-modern-portrait.html}{New York Times article} that discusses how artists use digital tools to imagine what George Washington would have looked like had he lived today).

In summary, our key contributions are:
\begin{itemize}
    \item \datasetname{}, a large, diverse and high-quality dataset that can serve a variety of computer vision and graphics tasks,
    \item a new task---transforming faces across time---and a method for performing this task that uses unique vectorial transformations to modify generator weights across different models,
    \item and quantitative and qualitative results that demonstrate that our method can successfully transform faces across time.
\end{itemize}

\ignorethis{
Prior work have demonstrated that manipulation of real images can be achieved by encoding them into this space and learning a desirable transformation of the latent code, while keeping the generator fixed (cite). Inspired by these works, we propose a neural projection-based framework that projects a latent code encoding a real image onto a set of \emph{learnable} basis vectors spanning a target decade. The projected code is then fed to our time-spanning generator to output an image of the input identity, as if it were captured in the target decade.}
\section{Related Work}
\label{sec:related}
\subsection{Image analysis across time}
While common image datasets such as ImageNet \cite{imagenet} and COCO \cite{coco} do not explicitly use time as an attribute, 
those that do
show unique characteristics. Here we focus on image datasets that feature people.

\smallskip
\noindent\textbf{Datasets.} The Yearbook Dataset \cite{ginosar2015century} is a collection of 37,921 front-facing portraits of American high school students from the 1900s to the 2010s. The authors design models to predict when a portrait is taken. They also analyze the prevalence of smiles, glasses, and hairstyles across different eras.
The IMAGO Dataset \cite{IMAGO} contains over 80,000 family photos from 1845 to 2009. Images are labeled in ``socio-historical classes'' such as \textit{free-time} or \textit{fashion}. 
Hsiao and Grauman \cite{cultureclothing2021} collect news articles and vintage photos and build a dataset that feature fashion trends in the 20th century. 
Our dataset contains portraits from a much wider age range (the Yearbook Dataset focuses on high school students), from diverse geographic areas (the IMAGO Dataset focuses on Italian families), and exhibiting rich variation in occupation and styles (not just fashion images). 

\smallskip
\noindent\textbf{Analysis.} 
A standard task applicable to images with temporal information is to predict when they were taken, i.e., the date estimation problem. 
Müller-Budack et al. \cite{whenwastaken} train two GoogLeNet models, one for classification and one for regression to predict the date of a photo. 
Salem et al.~\cite{salem2016face2year} train different CNNs for date estimation using the face, torso, and patch of a portrait image. 
Other rich information can also be learned from temporal image collections.
In StreetStyle and GeoStyle~\cite{matzen2017streetstyle,mall2019geostyle}, a worldwide set of images taken between 2013-2016 were analyzed to discover spatio-temporal fashion trends and events. 
In \cite{cultureclothing2021}, 
topic models and clusterings are used to discover trends in news and vintage photos. Additionally, in \cite{Lee2013StyleAwareMR}, a weakly-supervised approach is used to discover stylistic trends in cars over the decades. \cite{linking2015iccp} learn trends over two centuries of architecture using Street View images. \cite{MatzenECCV14,liu2020factorize} also disentangle pictures of buildings in a spatio-temporal way to see how structures have changed over time.
Unlike prior works that focus on \textit{analyzing} temporal characteristics in the data, we work on the more challenging task of \textit{modifying} such characteristics.

\subsection{Portrait editing}
Editing of face attributes has been extensively studied since before the deep learning era.
A classic approach is to fit a 3D morphable model \cite{blanz1999morphable,3dmm-review} to a face image and edit attributes in the morphable model space. Other methods that draw on classic vision approaches includes Transfiguring Portraits~\cite{kemelmacher2016transfiguring}, which can render portraits in different styles via image search, selection, alignment, and blending.
Given the recent success of StyleGAN (v1 \cite{karras2019style}, v2 \cite{karras2020analyzing}, and v3 \cite{StyleGANv3}) in high quality face synthesis and editing, many works focus on editing portrait images using pre-trained StyleGAN models. In these frameworks, a photo is mapped into a code in one of StyleGANs latent spaces~\cite{ganinversion}. Feeding the StyleGAN generator with a modified latent code yields a modified portrait~\cite{shen2020interfacegan,Wu2021StyleSpaceAD}.
To find the latent code of an input image, one can directly optimize the code so that the StyleGAN can reconstruct the inputs \cite{Abdal2019Image2StyleGANHT,wulff2020improving} or train a feed-forward network such as pSp \cite{richardson2021encoding} and e4e \cite{tov2021designing}that directly predicts the latent code.
We adopt an optimization-based procedure to obtain better facial details.

Once a latent code of an image is obtained, portrait image editing can be done in the latent space with a pre-trained StyleGAN generator.
Directions for change of viewpoint, aging, and lighting of faces can be found by PCA in the latent space \cite{harkonen2020ganspace}, or from facial attributes supervision \cite{shen2020interpreting}.
Shen and Zhou \cite{shen2021closed} find that editing directions are encoded in the generators' weights and can be obtained by eigen decomposition.
Collins et al. \cite{instyle} perform local editing of portraits by mixing layers from reference and target images.
Alternatively, the StyleGAN generator can also be modified for portrait editing.
StyleGAN-nada \cite{gal2021stylegannada} and StyleCLIP \cite{Patashnik_2021_ICCV} use CLIP \cite{Radford2021LearningTV} to guide editing on images with target attributes. 
Toonify \cite{Pinkney2020ResolutionDG} uses layer-swapping to obtain a new generator from models in different domains.
Similar to StyleAlign~\cite{Wu2021StyleAlignAA}, we obtain a family of generators by finetuning a common parent model on different decades. The style change of a face is achieved by obtaining the latent code of the input image and feeding it into a modified target StyleGAN generator using PTI \cite{roich2021pivotal}. Our method is conceptually simple and doesn't require exploring the latent space.

Age editing on portraits is related to our task since both involve modifying the temporal aspects of an image.
In the work of Or-El, et al.~\cite{or2020lifespan}, age is represented as a latent code and applied to the decoder network of the face generator. An identity loss is used to preserve identities across ages.
Alaluf et al.~\cite{alaluf2021matter} design an age encoder that takes a portrait and a target age, produces a style code modification on pSp-coded styles and generates a new portrait with the target age.
In contrast, our editing aims to change the decade a photo was taken without altering the subject's age. 

GANs can also be used to recolor historic photographs \cite{zhang2017real,deoldify,Luo-Rephotography-2021}. 
In particular, Time-Travel Rephotography~\cite{Luo-Rephotography-2021} 
uses a StyleGAN2 model to project historic photos into a latent space of modern high-resolution color photos with modern imaging characteristics. 
Rather than focusing solely on low-level characteristics like color, our method alters a diverse collection of visual attributes, such as facial hair and make-up styles. 
Moreover, our method can transform images across a wide range of decades, instead of learning binary transformations between ``old'' and ``modern'' as in \cite{Luo-Rephotography-2021}.

\section{The \datasetname{} Dataset}

Our \datasetname{} (\datasetshortname) dataset features 26,247 images of notable people from the 19th to 21st centuries, with roughly 1,900 images per decade on average. It is sourced from Wikimedia Commons (WC), a crowdsourced and open-licensed collection of 50M images.

We automatically curate data from WC to construct \datasetshortname (Figure \ref{fig:samples}) as follows:
(1) The ``People by name'' category on WC contains 407K distinct people identities. We query each identity's hierarchy of people-centric subcategories (similar to~\cite{whoswaldo}) and organize retrieved images by identity. 
(2) We use a Faster R-CNN model~\cite{renNIPS15fasterrcnn,jiang2017face,2017arXiv170804370R} trained on the WIDER Face dataset \cite{yang2016wider} as a face detector. For each detected face, 68 facial landmarks are found using the Deep Alignment Network~\cite{kowalski2017deep}, and alignments are applied as in the FFHQ dataset~\cite{karras2019style} given these landmarks.
(3) We devise a clustering method based on clique-finding in face similarity graphs to group faces by identities (see appendix). %
This resolves ambiguities in photos that feature multiple people.
(4) We gather additional samples without biographic information from the ``19th Century Portrait Photographs of Women'' and ``20th Century Portrait Photographs of Women'' categories.  These make up about 15\% of the dataset. 

We leverage image metadata, identity labels, and biographic information available in WC to further assist in balancing and filtering our data. 
We discard any photos without time labels or taken before 1880, and sample a subset of 3,000 faces each for the 2000s and 2010s decades (which tend to feature many more images than other decades) to maintain a roughly balanced distribution of images across decades.
For images where the identity is known, we further filter by only keeping images where the identity is between 18 and 80 years old (comparing the image timestamp and identity's birth year).
We also estimate face pose using Hopenet~\cite{Ruiz_2018_CVPR_Workshops} and remove images with yaw or pitch greater than $30$ degrees.
After these automated collection and filtering steps, we manually inspected the entire dataset and removed images with clearly incorrect dates, images that were not cropped properly, images that were duplicates of other identities, and images featuring objectionable content. This resulted in a removal of $6\%$ of the assembled data.

The total number of samples from each decade in our curated dataset, demographic and other biographic distributions, and further implementation details can be found in our supplementary material. We create train and test splits by randomly selecting 100 images per decade as a test set, with the remaining images used for training.
Samples from the dataset are shown in Figure \ref{fig:samples}.

As detailed in Section \ref{sec:related}, the Yearbook dataset \cite{ginosar2015century} is another dataset that spans multiple decades and could potentially allow for transforming faces across time. However, it is grayscale only, contains one age group, and critically, its images are low resolution ($186\times 171$). %
In the supplementary material, we show that high-quality synthesized images cannot be obtained using this dataset, further highlighting the benefit of our \datasetshortname dataset.

\section{Transforming Faces Across Time} \label{sec:method}

\begin{figure*}[t]
    \centering
    \jsubfig{\includegraphics[height=4.15cm]{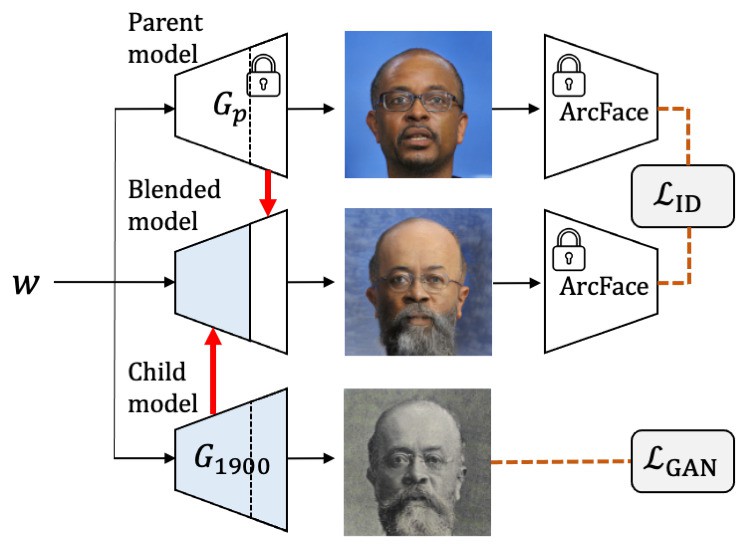}}{Learning Decade Models}
    \hspace{60px}
     \jsubfig{\includegraphics[height=4.15cm]{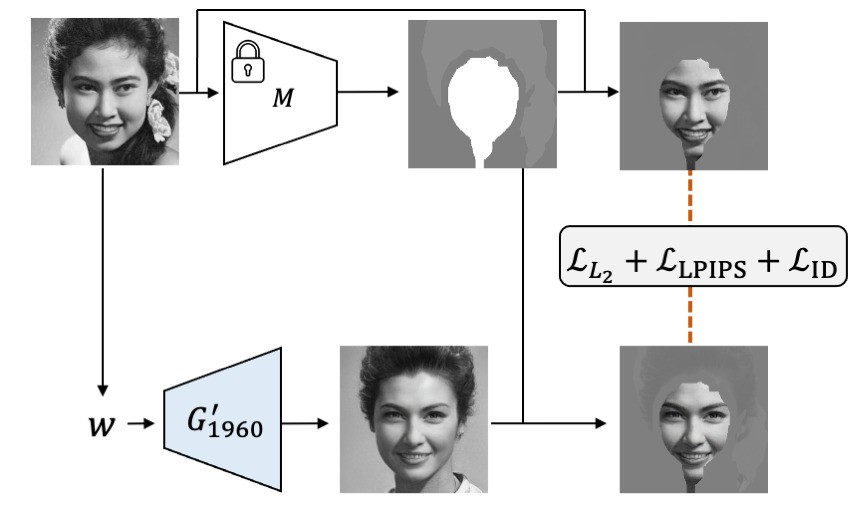}} {Single-image Refinement}
    \caption{\textbf{Overview of our method.} Left: We first train a family of StyleGAN models, one for each decade, using adversarial losses and an identity loss on a blended face, which resembles the parent model in its colors. %
    Right: Afterwards, each real image is projected onto a vector $w$ on the decade manifold (1960 in the example above). We learn a refined generator $G'_t$ and transfer the learned offset to all models (this process is visualized in Figure \ref{fig:pca}). To better encourage the refined model to preserve facial details, we mask the input image and apply all losses in a weighted manner (further described in the text).  
    }
    \label{fig:method}

\end{figure*}

Given a portrait image from a particular decade, our goal is to predict what the same person might look like across various decades ranging from 1880 to 2010.
The key challenges are: (1) maintaining the identity of the person across time, while (2) ensuring the result fits the natural distribution of images of the target decade in terms of style and other characteristics. 
We present a novel two-stage approach that addresses these challenges. Figure \ref{fig:method} shows an overview of our approach.

\begin{figure*}[t]
    \centering
    \includegraphics[width=1\textwidth]{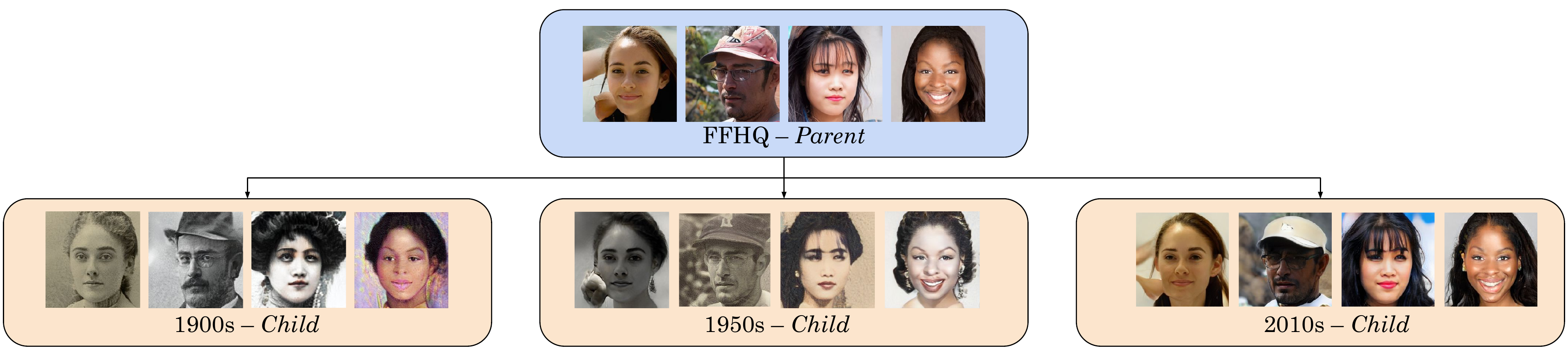}
    \caption{
    We finetune a family of decade generators (child models) from an FFHQ-trained parent model. While each generator captures unique styles, the generated images from the same latent code are aligned in terms of high-level properties such as pose.
    }
    \label{fig:model_family}
\end{figure*}

First, rather than training a single generative model that covers all decades  (e.g., \cite{or2020lifespan}), we train a family of StyleGAN models, one for each decade (Section \ref{sec:aligning}, Figure \ref{fig:model_family}). These are obtained by fine-tuning the same \emph{parent} model, resulting in a set of \emph{child} models whose latent spaces are roughly \emph{aligned} \cite{Wu2021StyleAlignAA}, as described in \Cref{sec:aligning}.
The alignment ensures that providing the same latent code to different models results in portraits with similar high-level properties, such as pose. %
At the same time, the resulting images exhibit the unique characteristics of each decade.
Given a real portrait from a particular decade, it is first inverted into the latent space of the corresponding model, and the resulting latent code can then be fed into the model of any desired target decade. Our approach makes it unnecessary to search for editing directions in the latent space (e.g., \cite{shen2020interfacegan,harkonen2020ganspace}).

Next, to better fit the identity of the input individual, we apply single-image finetuning of the family of per-decade StyleGAN generators (Section \Cref{sec:pti}). Specifically, we introduce Transferable Model Tuning (\methodshort{}), 
a modified PTI (Pivotal Tuning Inversion)~\cite{roich2021pivotal} procedure, to obtain an adjustment for the weights of the source decade generator, and apply the resulting adjustment to the target generator(s).
This input-specific adjustment is done in the generator's parameter space, enabling us to better preserve the input individual's identity, while maintaining the style and characteristics of the target decade. We now describe these two stages in more detail.

\subsection{Learning coarsely-aligned decade models}\label{sec:aligning}

We are interested in learning a family of StyleGAN2-ADA \cite{Karras2020ada} generators $\{G_{t}\},\; t \in \{1880,1890,\cdots,2010\}$ each of which maps a latent vector $w \in \mathcal{W}$ to an RGB image. 
For each decade, we finetune a separate StyleGAN model with weights initialized from an FFHQ-pretrained model. We call the FFHQ-pretrained model the \emph{parent model} $G_p$, and the finetuned network for decade $t$ the \emph{child model} $G_{t}$. Consistent with the findings in prior work \cite{Wu2021StyleAlignAA}, we observe that the collection of generators $\{G_{t}\},\; t \in \{1880,\cdots,2010\}$ exhibits semantic alignment of faces generated from the same latent code $w$: they share similar face poses and shapes. However, various fine facial characteristics such as eyes and noses, which are important for recognizing a person, often drift from one another (as evident in Figures~\ref{fig:model_family} and \ref{fig:pti_example}).

To better preserve identity across decades when finetuning each child model, to the standard StyleGAN objective $\mathcal{L}_{\text{GAN}}$ we add an identity loss. 
Specifically, we measure the cosine similarity between ArcFace \cite{deng2019arcface} embeddings of images generated by $G_p$ and by $G_{t}$:
\begin{equation}
    \mathcal{L}_{\text{ID}} = 1 -\texttt{cos\_sim}(\text{ArcFace}(G_p(w)), \text{ArcFace}(G^B_{t}(w))). 
\end{equation}
A similar loss is used in \cite{richardson2021encoding}. However, since the ArcFace model was only trained on modern day images, we found that this raw identity loss performed poorly on historical images, due to the domain gap. To solve this issue, we use a blended version $G^B_t(w)$ instead of the original $G_t(w)$. We create $G^B_t(w)$ using layer swapping \cite{Pinkney2020ResolutionDG} to mix $G_p(w)$ and $G_{t}(w)$ at different spatial resolutions:
we combine the \textit{coarse} layers of the child model $G_{t}(w)$ with the \textit{fine} layers of the parent model $G_p(w)$. By doing so, we ``condition'' our input image for the identity loss by making its colors more similar to the image generated by the parent, and thus more similar to the distribution of the images used to train the ArcFace model. 
Figure \ref{fig:method} (left) shows that the blended (middle) image retains the structure of the 1900s image, but its colors better resemble those of a modern day photo. In addition, this technique restricts the identity loss to focus on layers which generally control head shape, position, and identity. Note that this blended image is only used to compute the loss, and not in the transformed results.

\begin{figure*}[t]
\begin{minipage}{0.25\linewidth}
    \includegraphics[width=\linewidth]{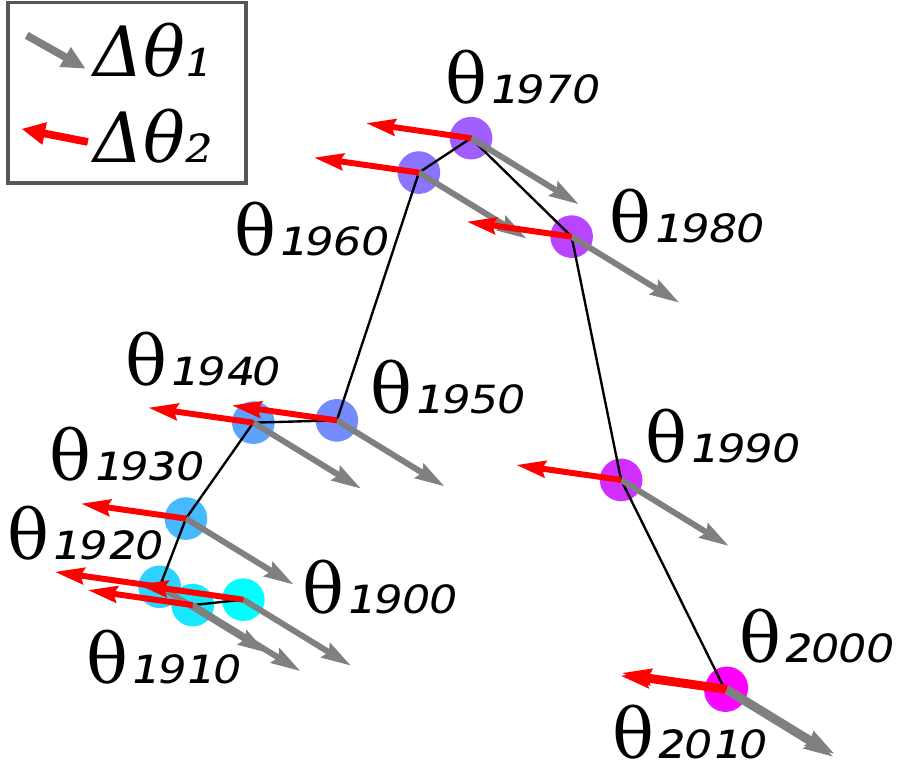}
\end{minipage}
\rotatebox[origin=c]{90}{\hspace{13pt}2 (1920)\hspace{20pt}1 (2010)}
\begin{minipage}{0.72\linewidth}
    \includegraphics[width=\linewidth]{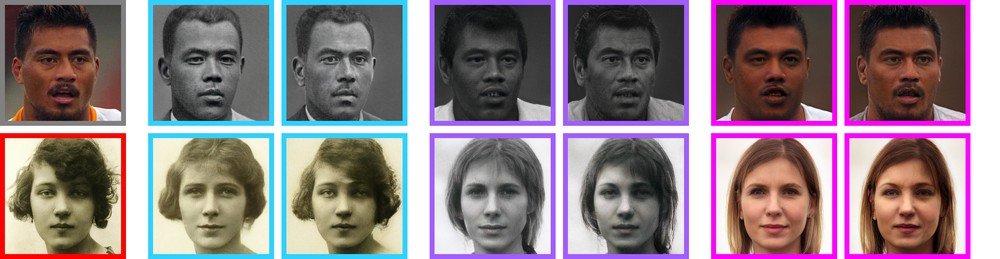}
    \begin{tabular}{m{33px}m{40px}m{42px}m{35px}m{42px}m{36px}m{42px}}
    \centering Input & \centering \hspace{3px}$\theta_{1920}$ & \centering $\theta_{1920} + \Delta \theta_i$ & \centering $\theta_{1970}$  & $\theta_{1970} + \Delta \theta_i$& \centering$\theta_{2010}$& $\theta_{2010} + \Delta \theta_i$
    \end{tabular}
\end{minipage}
    
    \caption{\textbf{Visualization of \methodshort{} offsets.} We obtain offset vectors $\Delta \theta_1$ (for the image in row 1) and $\Delta \theta_2$ (row 2) for the weights of the source decade generator and apply it to every target decade generator. On the left we use PCA to visualize the convolutional parameters for all target generators in 2D. Each dot represents the weights of a single generator, colored according to decade, and with edges connecting adjacent decades. We illustrate the offset vectors optimized for the two input images (colored in gray and red) and the corresponding transformed images for three different decades. For each decade we show images before (left) and after (right) applying \methodshort{}. Adding these offsets has the effect of improving identity preservation. 
    }
    \label{fig:pca}
\end{figure*}
\subsection{Single-image refinement and transferable model tuning}\label{sec:pti}
As previously described, we first train the family of aligned StyleGAN models with randomly sampled latent codes from $\mathcal{W}$ and our collections of real per-decade images. 
Given these coarsely-aligned per-decade models, we are given a single real face image as input and aim to generate a set of faces across various decades.
In order to better preserve the identity of the input image across these decades, we introduce Transferable Model Tuning (\methodshort{}). \methodshort{} is inspired by PTI~\cite{roich2021pivotal}, which
is a procedure for optimizing a model's parameters to better fit to an input image after GAN inversion.
\methodshort{} extends PTI from a single generator model to a family of models.
Our \methodshort{} procedure produces a set of face images where the identity is preserved in the presence of changing style over time (Figure \ref{fig:pca}).

Specifically, given an input image $x$ from decade $t$, we first obtain its latent code $w \in \mathcal{W}$ using the projection method from StyleGAN2: $w = \argmin_{w} ||x-G_t(w;\theta_t)||$, where $\theta_t \in \Theta$ is the vector of all parameters %
in $G_t$. We only work with child models in this stage.
As the first step of \methodshort{}, we fix the obtained latent code $w$ and optimize over $\theta_t$ to obtain a new model $G'_t $ with parameters $\theta'_t$ 
(Figure \ref{fig:method}, right).
\begin{equation}
    \theta'_t = \argmin_{\theta_t} ||x-G_t(w;\theta_t)||.
\end{equation}
Tuning in the parameter space $\Theta$ of generators, instead of only working in the latent space $\mathcal{W}$ as in previous work \cite{Nitzan2020FaceID}, allows us to better fit the facial details of the input individual, such as eyes and expression.
Treating $\theta_t$ and $\theta'_t$ as vectors in the parameter space $\Theta$, this tuning can be thought of as applying an offset $\Delta \theta = \theta'_t-\theta_t$ to the original model.

We found that this $\Delta \theta$ offset is surprisingly transferable from one \methodshort{}-tuned generator to all other decade generators in the parameter space $\Theta$. Concretely, to obtain the style-transformed face of $x$ in \emph{any} target decade $d$, we simply apply:
\begin{equation}
    x_d = G_d(w; \theta_d + \Delta \theta),
\end{equation}
where $G_d$ is the generator for decade $d$ and $\theta_d$ is the vector of all its parameters. We visualize the learned \methodshort{} offsets for two different input images in Figure \ref{fig:pca}. As illustrated in the figure, these offsets greatly improve the identity preservation of synthesized portraits. 

Intuitively, $\Delta \theta$ ``refines'' the parameters of the generator family to the single input face image to reconstruct better facial details. 
We hypothesize that the found $\Delta \theta$ offsets mainly focus on improving identities. Since the coarsely-aligned family of generators share similar weights that are responsible for identity-related features, those offsets are easily transferable.
 While most prior works, e.g., \cite{shen2020interfacegan,harkonen2020ganspace,Wu2021StyleSpaceAD} modify images using linear directions in various \textit{latent spaces} of StyleGAN, we are the first to apply a linear offset in the generator \textit{parameter space}, to a collection of generators.
 We hope our work will inspire future investigations on understanding the linear properties in GAN's parameter spaces. 
An analysis of the effects of applying \methodshort{} to a family of models can be found in the supplementary material.

As demonstrated in Figure \ref{fig:method} (right), to focus the loss computation on facial details instead of hair and background, 
we apply masks to images before calculating the losses.  We use a DeepLab segmentation network \cite{deeplabv3plus2018} trained on CelebAMask-HQ photos \cite{or2020lifespan,CelebAMask-HQ}. Empirically we determine it is best to apply a weight of 1.0 on the face, 0.1 on the hair and 0.0 elsewhere. We put a small weight on the hair to accurately reconstruct it, as it does contribute to the image's stylization. However, we do not want to prioritize it over facial features. In addition, we find that it is best to keep StyleGAN's ToRGB layers frozen. Otherwise, color artifacts are introduced into the generators. We follow the objectives introduced in \cite{roich2021pivotal} and minimize a perceptual loss ($\mathcal{L}_{\text{LPIPS}}$), and a reconstruction loss ($\mathcal{L}_{\text{L2}}$). In addition, we add another identity loss ($\mathcal{L}_{\text{id}}$) to further enhance identity preservation for the generated images.

Using our two-stage approach, for each portrait, we obtain a set of faces that maintain the identity as well as demonstrate diverse styles across decades.

\section{Results and Evaluation}
\label{sec:results}
We conduct extensive experiments on the \datasetname{} test set. We compare our approach to several state-of-the-art techniques across a variety of metrics that quantify how well transformed images depict their associated target decades and to what extent the identity is preserved. 
We also present an ablation study to examine the impact of the different components of our approach. 
Additional uncurated results and visualizations of the full test set (1,400 individuals) are in the supplementary material.

\subsection{State-of-the-art Image Editing Alternatives}
As no prior works directly address our task, 
we adapt commonly used image editing models to our setting and perform comprehensive comparisons.

\medskip \noindent \textbf{Image-to-image translation.} Unpaired image-to-image translation models learn a mapping between two or more domains. We train a StarGAN v2~\cite{Choi2020StarGANVD} model on our dataset, where decades are domains. Because the quantitative metrics between our model and StarGAN are so similar, we show a detailed qualitative comparison between our model and StarGAN in the supplementary material, over a set of individuals balanced by ethnicity and gender. We see that StarGAN has a poor understanding of skin tone and overall identity, which is critical for our real-world applications. Results on another model, DRIT++~\cite{Lee2020DRITDI}, are in the supplementary material.

\medskip \noindent \textbf{Attribute-based editing.} 
We consider each decade as a facial attribute and compare against recent works performing attribute-based edits.
While many attributes in prior work are binary 
\cite{shen2020interfacegan,harkonen2020ganspace}, our decade attribute has multiple classes. By comparison, age-based transformation is more similar to our problem setting, as age is often broken up into $K$ bins~\cite{or2020lifespan}. We compare against SAM~\cite{alaluf2021matter}, a recent age-based transformation technique that also operates in the StyleGAN space.

\medskip \noindent \textbf{Language-guided editing.} Weakly supervised methods that leverage powerful image-text representations (\emph{e.g.} CLIP~\cite{Radford2021LearningTV}) have demonstrated impressive performance in portraits editing. We compare against: (i) StyleCLIP~\cite{Patashnik_2021_ICCV}, which learns latent directions in StyleGAN's $\mathcal{W+}$ space for a given text prompt, and (ii) StyleGAN-nada (StyleNADA)~\cite{gal2021stylegannada}, which modifies the weights of the StyleGAN generator based on textual inputs. For both models, we used the text prompt ``A person from the [XYZ]0s'', where [XYZ]0 is one of \datasetshortname's 14 decades (1880-2010). 
For StyleCLIP, we use models trained on both FFHQ and \datasetshortname. 
Because StyleGAN-nada is designed for out-of-domain changes, we experiment with how well it can modify the generator from the FFHQ space to various decades in our dataset. 
We use 100 FFHQ photos. We compare these two baselines to how well our model can transform FFHQ images to each of the 14 decades. 

\medskip \noindent \textbf{Conditioning on a single model vs. multiple models.} We train one model for each decade as done in StyleAlign~\cite{Wu2021StyleAlignAA} because it is difficult to disentangle ecah decade's style in a single model. The modifications we made to SAM~\cite{alaluf2021matter}, such as adding a decade classifier, can be considered a straightforward way to condition decade labels on a single model's latent space. In addition, StyleCLIP~\cite{Patashnik_2021_ICCV} also uses a single finetuned model to transform images within a GAN's latent space. Our results show that these single-model methods struggle to understand the style of each decade, as compared to our multiple-model approach.

\medskip \noindent \textbf{Time Travel Rephotography.} Similar to our method, Time Travel Rephotography \cite{Luo-Rephotography-2021} is designed to imagine what a historical person would have looked like in another time period (and can only perform a transformation to modern imagery). However, the authors focus primarily on image restoration as opposed to changing a person's style or fashion. We see that Time Travel Rephotography improves camera quality and lighting instead of changing hairstyles and other stylistic features as our model does. Technically, \cite{Luo-Rephotography-2021} inverts their photos into a pretrained StyleGAN on FFHQ instead of training new models on data for each decade. We provide visual comparisons between our model and Time Travel Rephotography in the supplementary material.

\begin{table}[t]
\setlength{\tabcolsep}{3.0pt}
 \def\arraystretch{1.1}
\centering

\resizebox{\linewidth}{!}{
\begin{tabular}{llccccccc}
\toprule
    &Method &FID$\downarrow$ & KMMD $\downarrow$& $\text{DCA}_0$$\uparrow$ & $\text{DCA}_1$$\uparrow$ & $\text{DCA}_2$$\uparrow$ & $\text{ID}_{acc}$$\uparrow$  \\ \midrule 
    \multirow{3}{*}{\rotatebox[origin=c]{90}{FFHQ}} & StyleCLIP & 254.39 & 1.87 & 0.08 & 0.18& 0.36 & \textbf{0.99}\\
    & StyleNADA & 312.06 & 2.03 & 0.10 & 0.30 & 0.38&  0.96 \\
    & Ours &\textbf{69.46}& \textbf{0.43}& \textbf{0.50} & \textbf{0.81} & \textbf{0.91} & 0.93  \\
    \midrule 
    \multirow{4}{*}{\rotatebox[origin=c]{90}{\datasetshortname}} & StarGAN v2 & 68.05 & \textbf{0.40} &  0.38 & 0.75 & 0.89 & 0.97\\
    & SAM & 96.52 & 0.72 &  \textbf{0.51*} & \textbf{0.85}* & 0.89* & \textbf{1.00} \\
    & StyleCLIP & 108.25 & 0.85 & 0.07 & 0.21 & 0.36 & \textbf{1.00}\\
    & Ours & \textbf{66.98} &\textbf{0.40} &  \textbf{0.47} & \textbf{0.78} & \textbf{0.90}&0.99 \\
\bottomrule
\end{tabular}
}
\caption{\textbf{Quantitative Evaluation.} We compare performance against SOTA techniques on FFHQ (top three rows) and on our test set (bottom four rows). Our method outperforms others in terms of most metrics. *Note that SAM uses the decade classifier during training and therefore the DCA metric is skewed in this case, as we further detail in the text. 
}
\label{tab:baseline-stats}
\end{table}

\begin{figure*}
 \raggedright
\jsubfig{\includegraphics[width=0.48\linewidth]{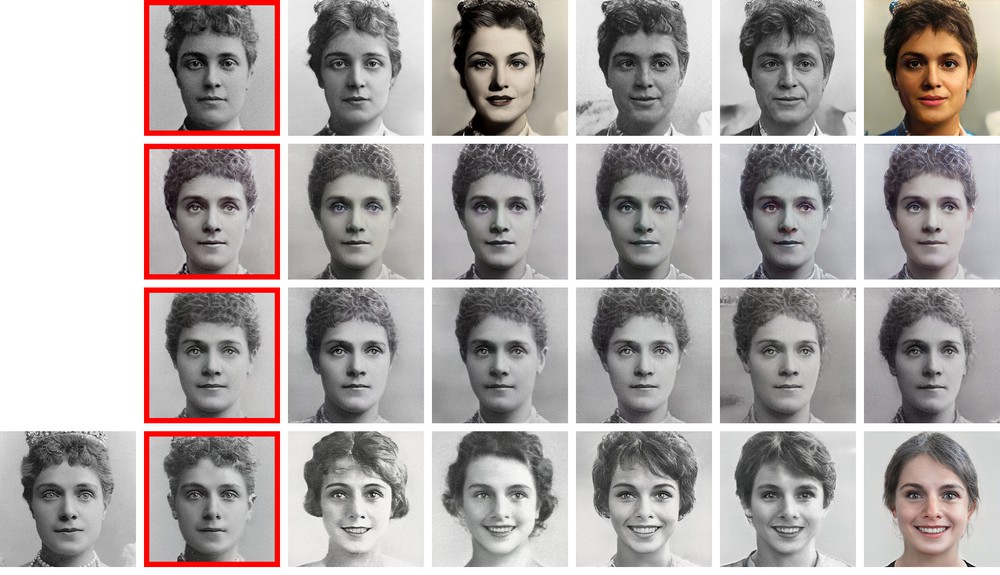}}{} \hfill
\jsubfig{\includegraphics[width=0.48\linewidth]{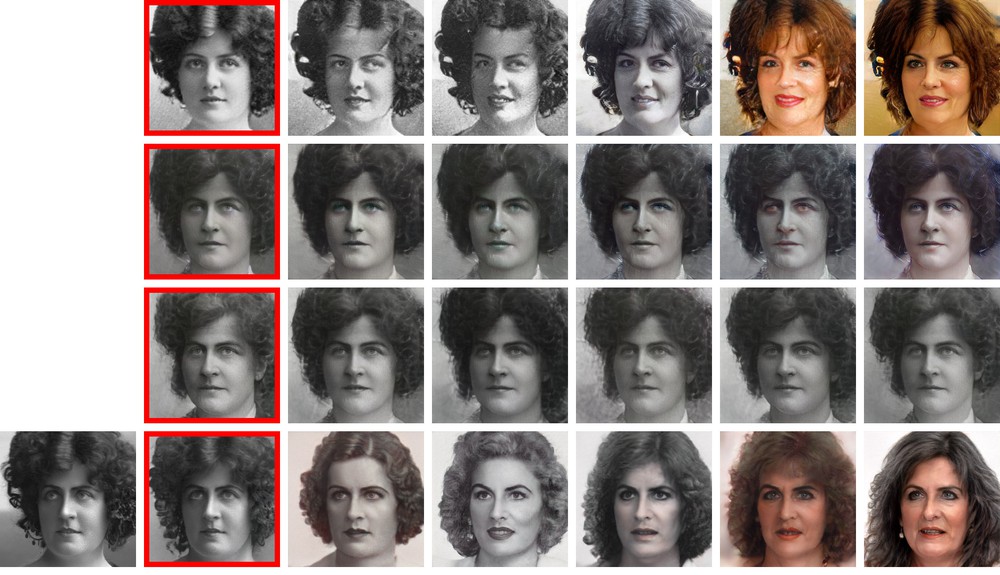}}{}
\rotatebox[origin=rB]{-90}{\begin{tabularx}{140px}{FFFF}\tiny{StarGAN} & SAM & \tiny{StyleCLIP} & Ours \end{tabularx}}

\jsubfig{\includegraphics[width=0.48\linewidth]{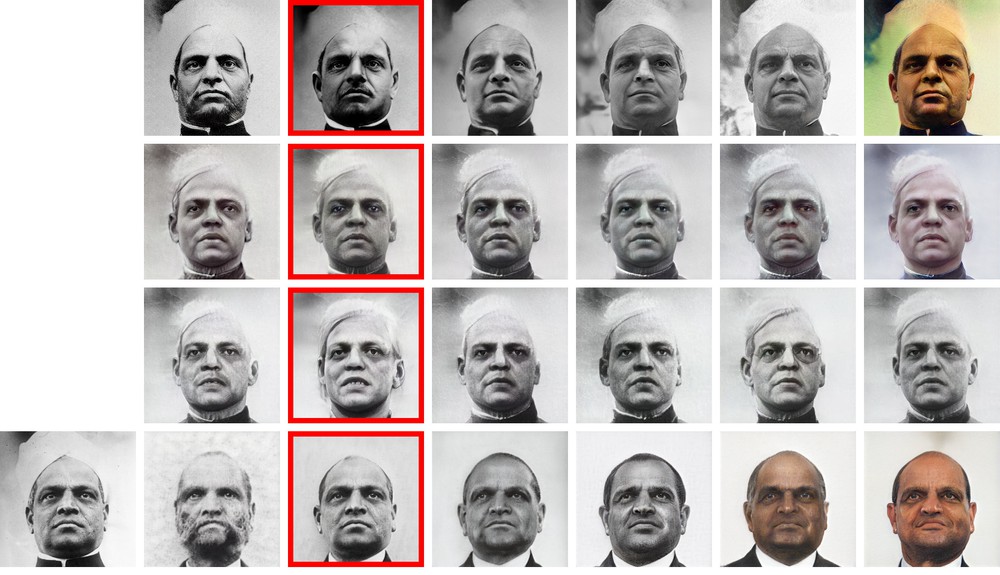}}{} \hfill
\jsubfig{\includegraphics[width=0.48\linewidth]{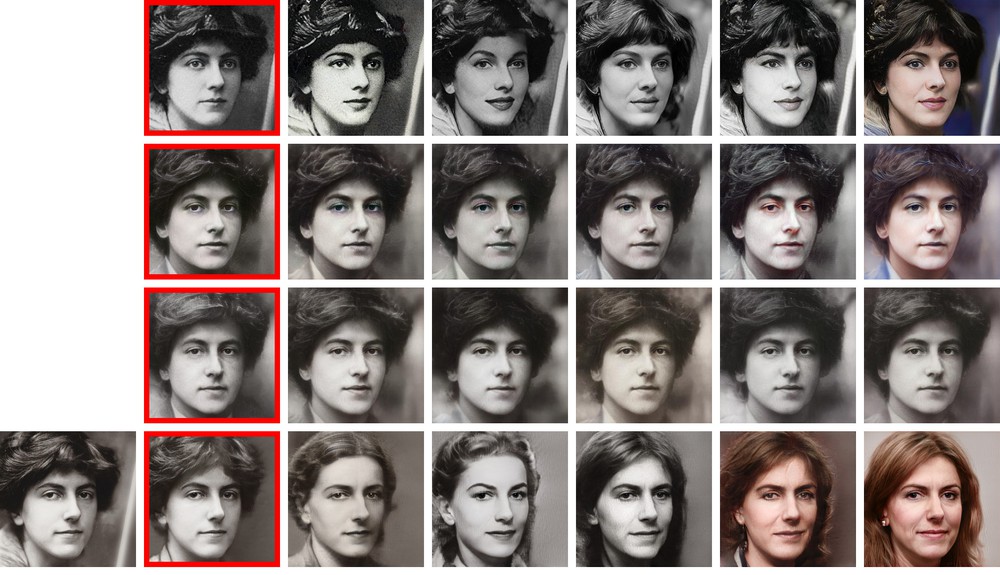}}{}
\rotatebox[origin=rB]{-90}{\begin{tabularx}{140px}{FFFF}\tiny{StarGAN} & SAM & \tiny{StyleCLIP} & Ours \end{tabularx}}

\jsubfig{\includegraphics[width=0.48\linewidth]{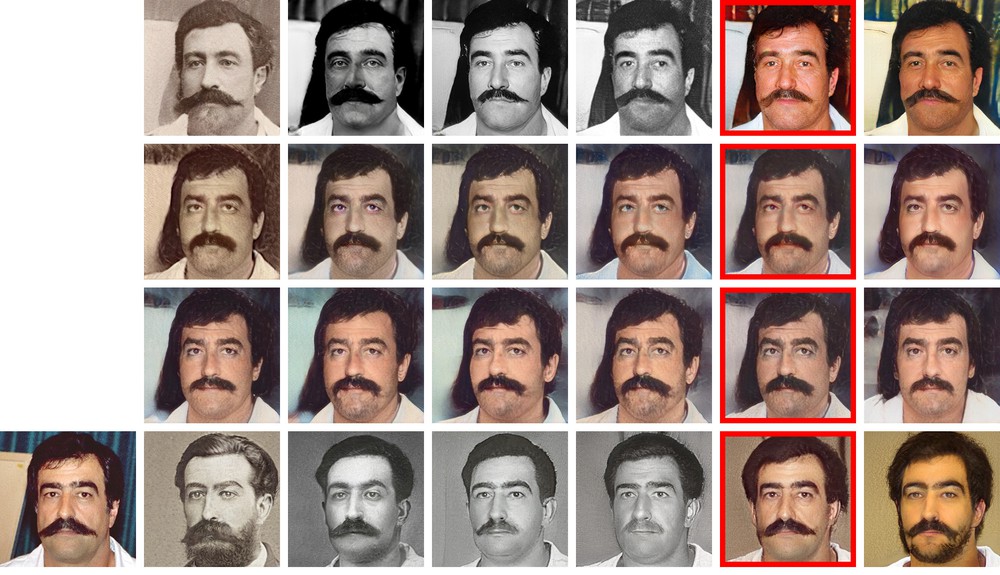}}{} \hfill
\jsubfig{\includegraphics[width=0.48\linewidth]{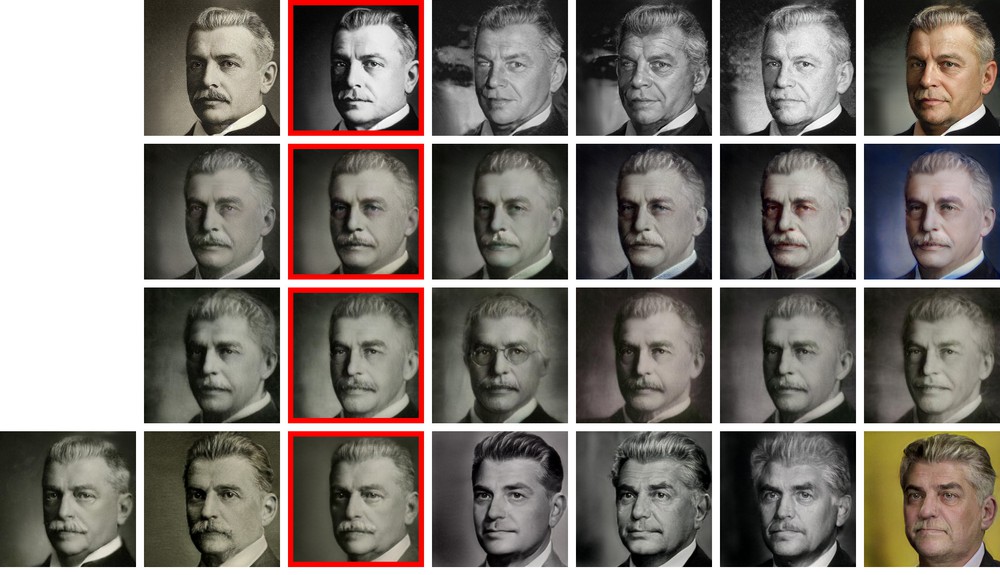}}{}
\rotatebox[origin=rB]{-90}{\begin{tabularx}{140px}{FFFF}\tiny{StarGAN} & SAM & \tiny{StyleCLIP} & Ours \end{tabularx}}

\jsubfig{\includegraphics[width=0.48\linewidth]{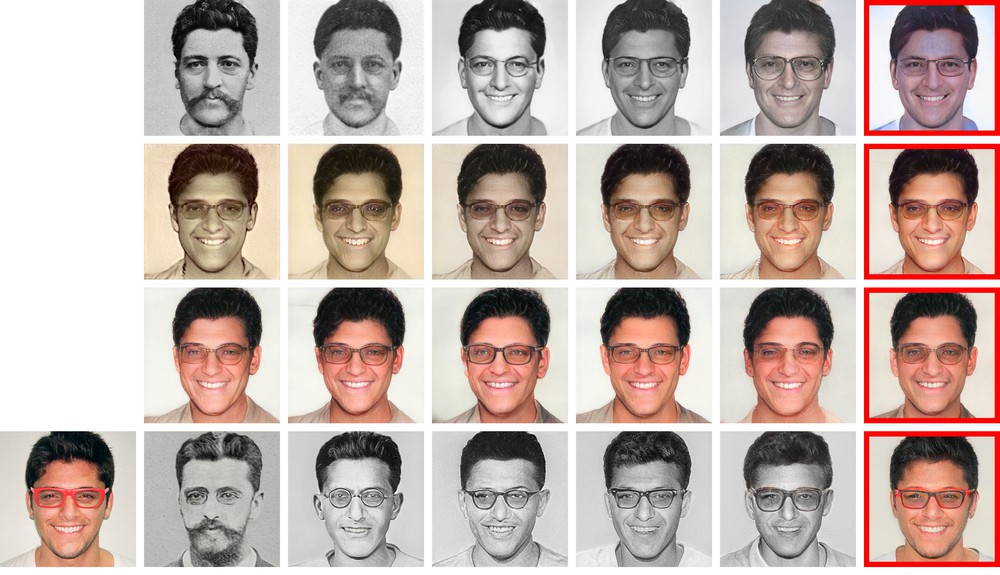}}{} \hfill
\jsubfig{\includegraphics[width=0.48\linewidth]{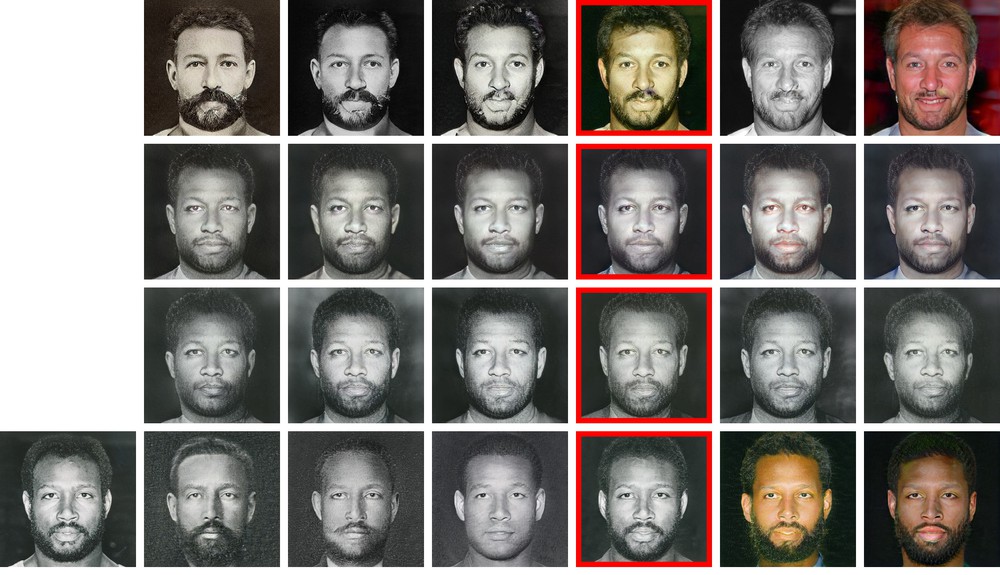}}{}
\rotatebox[origin=rB]{-90}{\begin{tabularx}{140px}{FFFF}\tiny{StarGAN} & SAM & \tiny{StyleCLIP} & Ours \end{tabularx}}

\begin{tabularx}{0.48\linewidth}{FFFFFFF}
Input & 1880s & 1920s & 1940s & 1960s & 1980s  & 2010s
\end{tabularx} \hfill
\begin{tabularx}{0.48\linewidth}{FFFFFFF}
Input & 1910s & 1930s & 1950s & 1970s & 1990s & 2010s
\end{tabularx} \hspace{11px}
    \caption{\textbf{Qualitative Results}. Above we compare results  generated by baselines and our technique. The red box indicates the inversion of the original input. We observe that our approach allows for significant changes across time while best preserving the input identity.  While SAM and StarGAN are able to stylize images, these changes are mostly limited to color. StyleCLIP struggles to generate meaningful changes. Please refer to the supplementary material for qualitative results on the full test set.}
    \label{fig:qualitative_fwt}
\end{figure*}

\subsection{Metrics}
\noindent\textbf{Visual quality}. We use the standard FID \cite{Heusel2017GANsTB,Seitzer2020FID} metric as well as the Kernel Mean Maximum Discrepancy Distance (KMMD)~\cite{wang2020minegan} metric. 
As image quality varies across decades, we compute scores between real and edited portraits separately for each decade and then average over all decades. 
Because FID can be unstable on smaller datasets~\cite{chong2019effectively}, similar to prior work~\cite{noguchi2019image,wang2020minegan}, we measure KMMD ~\cite{wang2020minegan} on Inception~\cite{Szegedy2016RethinkingTI} features. Experimentally, we find that these two scores are highly sensitive to an image's background. Therefore, we compute the scores on images of size $256\times 256$ cropped to 160$\times$160 pixels.

\noindent\textbf{Decade style}. We evaluate how well the generated samples capture the style of the target decade using a EfficientNetB0 classifier \cite{Tan2019EfficientNetRM} that we trained separately. Using the classifier, we define the Decade Classification Accuracy ($\text{DCA}$). 
We follow prior works \cite{gaudette2009evaluation} and report three metrics: $\text{DCA}_0$, $\text{DCA}_1$ and $\text{DCA}_2$,
where $\text{DCA}_p$ measure the accuracy within a tolerance of $p$ decades.

\noindent\textbf{Identity preservation}. 
We use the Amazon Rekognition service to measure how well a person's identity has been preserved in generated portraits.
Their \href{https://docs.aws.amazon.com/rekognition/}{\textsc{CompareFaces}}
operation outputs a similarity score between two faces. 
As a metric, we report $\text{ID}_{acc}$  -- the fraction of successful identity comparisons. We consider a comparison to be successful if its similarity score is above a certain threshold (set empirically to $1.0$). 

\begin{figure}[t]
\centering
\jsubfig{\includegraphics[width=\linewidth]{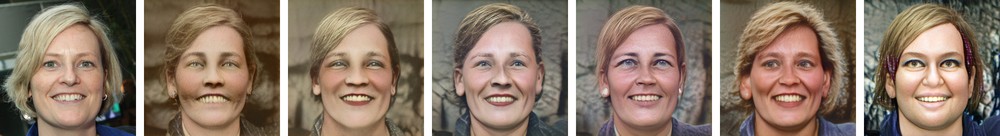}}{}
\jsubfig{\includegraphics[width=\linewidth]{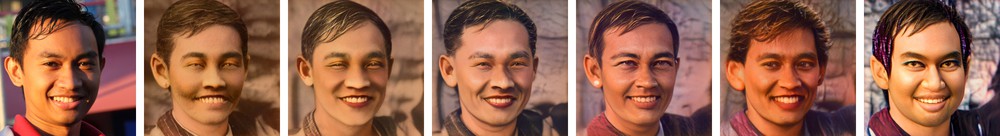}}{}
\begin{tabularx}{\linewidth}{FFFFFFF}
Input  & 1900s & 1920s & 1940s & 1960s  & 1980s &  2010s
\end{tabularx}
    \caption{\textbf{StyleGAN-nada Results}. 
    Although StyleGAN-nada~\cite{gal2021stylegannada} produces some style changes across decades, all images are transformed similarly. For example, all 1980s portraits adopt a frizzy hairstyle.
    }\label{fig:qualitative_nada}
\end{figure}
\subsection{Results}
We present test set performance for all methods in Table \ref{tab:baseline-stats} and qualitative results in Figure \ref{fig:qualitative_fwt}. Results on the \emph{full} test set (1400 samples) are provided using an interactive viewer as part of our supplementary material. We present additional results on StyleGAN-nada's effect on FFHQ input images in Figure \ref{fig:qualitative_nada}. We see that our method performs well numerically in terms of all metrics. Although StarGAN also performs well in FID and KMMD, the style changes by StarGAN are mostly about color; there are few changes to makeup, hair, and beards, whereas our model can perform such changes. Our method also has fewer artifacts. As illustrated in Figure \ref{fig:qualitative_nada}, modifications from StyleGAN-nada are more caricature-like than realistic. As a result, StyleGAN-nada performs poorly with respect to FID, KMMD, and DCA metrics. 
 For StyleCLIP and SAM, the identity preservation is near 1.0 because their changes are generally limited to color. Nonetheless, our model still successfully matches input and transformed images in most cases (for $93\%$ and $99\%$ of samples, for FFHQ and \datasetshortname{} images, respectively), while generating significant changes in styles. While SAM performs the best with respect to DCA, we believe this is advantaged because SAM used the same classifier during training as a loss function. In fact, in Figure~\ref{fig:qualitative_fwt}, SAM demonstrates little change across decades. We suspect that the classifier is leading SAM to overfit to noise, instead of truly changing an image's style.

From our results we can discern interesting details that  provide insight into style trends across time. 
For example, as illustrated in he top left example in Figure \ref{fig:qualitative_fwt}, we see that the individual adopts a bob haircut, one popularized by \href{https://historydaily.org/the-bob-a-revolutionary-and-empowering-hairstyle}{Irene Castle}, and strongly associated with the flapper culture of the Roaring Twenties. Later on, we notice more contemporary hair styles. Finally, in the 2010s, we see that women tend to adopt longer hairstyles. 
Despite these generalizations, the portraits remain well-conditioned on the input, reflecting an individual's identity and aspects of their personal style. For instance, the individual on the left with the long mustache maintains facial hair across the decade transformations, although in very different styles. In the bottom left example, we observe glasses that change style over time. Not only does our model generate realistic transformations, but also captures the nostalgia of various time periods.

\begin{figure}[t]
\vspace{-4pt}
\raggedright

\jsubfig{\includegraphics[width=0.945\linewidth]{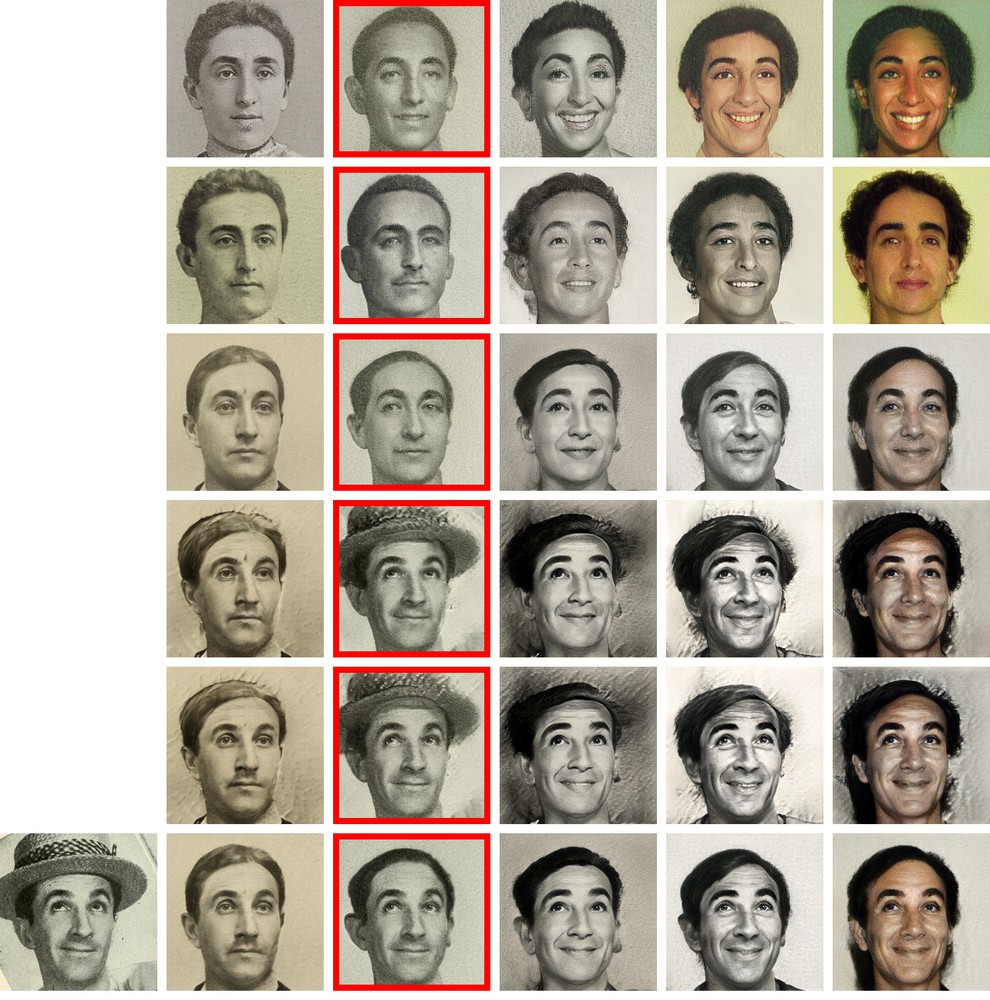}}{}
\rotatebox[origin=rB]{-90}{\begin{tabularx}{230px}{FFFFFF}\tiny{No $\mathcal{L}_{\text{id}}^{\text{(I)}}$} & \tiny{No LS} & \tiny{No TMT} &\tiny{No $\mathcal{L}_{\text{id}}^{\text{(II)}}$} & \tiny{No Mask}& \tiny{Ours} \end{tabularx}}
\begin{tabularx}{0.945\linewidth}{FFFFFF}
Input & 1890s & 1910s & 1930s & 1950s & 1970s
\end{tabularx}
\caption{\textbf{Ablations}. We see significant improvement in terms of identity preservation after adding the identity loss and \methodshort{} during training. The masking procedure alleviates artifacts caused by other regions in the image (such as the hat in the example above) by focusing the model's attention on facial details and allowing for larger modifications in other regions. See Table~\ref{tab:ablations} for descriptions of the ablation labels.}
  \label{fig:pti_example}
\end{figure}

\subsection{Ablations}
We present ablations on components in our approach in Table \ref{tab:ablations} and Figure \ref{fig:pti_example}. Specifically, we train five ablated models: (1) without the identity loss (for learning decade models), (2) without the blended image obtained using layer swapping, (3) without \methodshort{}, (4) without the identity loss (during \methodshort{}), and (5) without masking the images. The first ablation is akin to finding the nearest neighbor of an image in the StyleGAN $\mathcal{W}$ space using latent projection. As shown in the first row of the table, our baseline method already captures a decade's style well. However, the images are not aligned with respect to an input's identity, which is reflected in its low $\text{ID}_{acc}$ score. As a result, $\mathcal{L}_\text{id}^{\text{(I)}}$, $\mathcal{L}_\text{id}^{\text{(II)}}$, and \methodshort{} are necessary for identity preservation. %
In addition, we find that using masks during \methodshort{} reduces artifacts in generated images as it allows for an accurate inversion in the facial region and larger modifications in other regions in the image. 
Empirically, we notice that this reduces noise and improves decade classification. While FID, KMMD and DCA scores remain similar across the ablations, our full model shows strong improvement in $\text{ID}_{acc}$, which is the main objective of $\mathcal{L}_\text{id}^{\text{(I)}} $ and our proposed \methodshort{} stage. 
We also experimentally find that $\mathcal{W+}$ spaces \cite{richardson2021encoding} are less well aligned than the $\mathcal{W}$ space. More details are in the supplementary material. 
\begin{table}[h]
\centering
\resizebox{\linewidth}{!}{
\begin{tabular}{lllll|cccccc}
\hline

$\mathcal{L}_\text{id}^{\text{(I)}}$  & LS&  TMT& $\mathcal{L}_\text{id}^{\text{(II)}}$ & \small{Mask} & FID & KMMD & $\text{DCA}_0$  & $\text{DCA}_1$ & $\text{DCA}_2$ & $\text{ID}_\mathit{acc} $\\ 
\hline
$\times$& $\times$&$\times$& $\times$ & $\times$ & 69.51 &0.45&0.49 & 0.79 & \textbf{0.92} &0.61    \\ 
\checkmark& $\times$&$\times$& $\times$ &$\times$ & 69.36 &0.47& \textbf{0.51} &\textbf{0.82} & \textbf{0.92} & 0.63   \\ 
\checkmark& \checkmark	& $\times$	& $\times$	& $\times$ & 68.18  & 0.45 & 0.50  & \textbf{0.82} & \textbf{0.92} & 0.72    \\ 
\checkmark& \checkmark	& \checkmark	& $\times$	& $\times$& 67.32 & 0.39 & 0.46 & 0.78 & 0.89 & 0.95   \\ 
\checkmark& \checkmark& \checkmark	& \checkmark & $\times$ & 67.08 & \textbf{0.38} & 0.46 & 0.77 & 0.89   & \textbf{0.99} \\ 
 \checkmark& \checkmark	& \checkmark	& \checkmark & \checkmark & \textbf{66.98} & 0.40  & 0.47 & 0.78 & 0.90  & \textbf{0.99}   \\ 

\hline

\end{tabular}
}
\caption{\textbf{Ablation study} evaluating the effect of the identity loss while learning decade models ($\mathcal{L}_\text{id}^{\text{(I)}}$) and during \methodshort{} ($\mathcal{L}_\text{id}^{\text{(II)}}$), using a blended image with layer swapping (LS), \methodshort{}, and masking the images during \methodshort{} (Mask).  %
}
\label{tab:ablations}
\end{table}

\ignorethis{\medskip \noindent \textbf{FID} FID \cite{Heusel2017GANsTB, Seitzer2020FID} is an established way to measure the realism of GAN-generated images. For each input decade and target decade pair, we compared the FID between 100 real images from the target decade and 100 generated images.
 
\medskip \noindent \textbf{Classifier} Even for humans, determining what decade an image comes from is a difficult problem. To measure historical accuracy, we trained a ResNet-18 model on our dataset to classify an image's decade. Using the classifier, we adopted two metrics: \textit{ACC0} and \textit{ACC1} from \cite{gaudette2009evaluation}. \textit{ACC1} measures the classification accuracy within a tolerance of 1 decade, while \textit{ACC0} measures the traditional accuracy. When testing on unseen WikiTime images, we were able to achieve $50.25 \%$ on \textit{ACC0} and $78.67\%$ on \textit{ACC1} as seen in Figure \ref{fig:classifier-confusion}.
\hadar{The text and notations need to be revised according to the conventions used here \cite{gaudette2009evaluation}. Let's also report ACC2 (in addition to ACC0 and ACC1) }

\medskip \noindent \textbf{Identity} We measured how well each transformation preserves the identity of an input image using the Amazon Rekognition face verification API. \hadar{Human evaluation? In SAM, they transformed images of celebrities and asked workers to identify the transformed images.

}
\begin{enumerate}
    \item Show changes with only mean and standard deviation
    \item Fixed basis vectors with least squares optimization (PCA)
    \item Ablating the architecture
    \begin{enumerate}
        \item Fixed basis vectors with network 
        \item Without a projection layer 
    \end{enumerate}
    \item Ablating the losses
\begin{enumerate}
    \item Replacing adversarial loss with classification loss, showing that the GAN is needed
    \item Adding identity loss (and cyclic loss?) to "coarse" model, showing that the 2 model approach is needed
\end{enumerate}
\end{enumerate}
}

\section{Ethical Discussion}\label{sec:ethical}
Face datasets---and the tasks that they enable, such as face recognition---have been subject to increasing scrutiny and recent work has shed light on the potential harms of such data and tasks~\cite{bias1,bias2}. 
With awareness of these issues, our dataset was constructed with attention to ethical questions. The images in our dataset are freely-licensed and provided through a public catalog. We will include the source and license for 
each image in our dataset. As part of our terms of use, we will only provide our dataset for academic use under a restrictive license. Furthermore, our dataset does not contain identity information (and only includes one face per identity), and therefore cannot readily be used for facial recognition. Nonetheless, our dataset does inherit biases that are present in Wikimedia Commons. For instance, the data is gender imbalanced, containing a ratio of roughly $2:1$ male to female samples (according to the binary gender labels available on Wikimedia Commons, which are annotated by Wikipedia contributors). 
While such biases can be mitigated by balancing the data for training and evaluation purposes, we plan to continue gathering more diverse data to address this underlying bias in the data. For additional details on various features of our dataset, please refer to the accompanying datasheet~\cite{gebru2021datasheets}.

There are also ethical considerations relating to the risks of using portrait editing for misinformation.
Our task is perhaps less sensitive in this regard, since our explicit goal is to create fanciful imagery that is clearly anachronistic. 
That said, any results from such technology should be clearly labeled as imagery that has been modified.%

\section{Conclusion}
We present a dataset and method for transforming portrait images across time. 
Our \datasetname{} dataset spans diverse geographical areas, age groups, and styles, allowing one to capture the essence of each decade via generative models. 
By learning a family of generators and efficient tuning offsets, our two-stage approach allows for significant style changes in portraits, while still preserving the appearance of the input identity.
Our evaluation shows that our approach outperforms state-of-the-art face editing methods.
It also reveals interesting style trends existing in various decades.
However, our method still has limitations. As with any data-driven technique, our results are affected by biases that exist in the data. For instance, females with short hair are less common at the beginning of the 20th century, which may yield gender inconsistencies when transforming a short-haired modern female face to these early decades, including unexpected changes in visual features often associated with gender. %
In the future, we plan to explore methods that can improve consistency, perhaps by devising a way to jointly optimize models for different decades that better enforces consistency among them.
Finally, we envision that future uses of our data could go beyond the synthesis tasks we consider in our work, and explore the combination of both analysis and synthesis.

\section{Acknowledgements}
This work was supported in part by the National Science Foundation (IIS-2008313).

\printbibliography

\clearpage
\appendix
{\LARGE\textbf{Appendix}}

In this document, we present implementation details and additional results that are supplement to the main paper.  Section~\ref{supp:dataset} reports details about our \datasetname dataset. Section~\ref{supp:results} provides details about the adaptation of state-of-the-art image editing alternatives as well as additional results and analysis. 
Section~\ref{supp:details} provides implementation details such as training procedures and hardware specifications.
A datasheet for our dataset is provided separately. Note that we also provide an interactive viewer that demonstrates our results, as well as those obtained using alternative techniques, on the full test set (as well as a lighter viewer that only displays results for 100 random test samples).

\section{\datasetname Dataset Details}\label{supp:dataset}

\begin{figure}[ht!]
\centering
\begin{minipage}{0.49\textwidth}
\centering
\includegraphics[width=\textwidth]{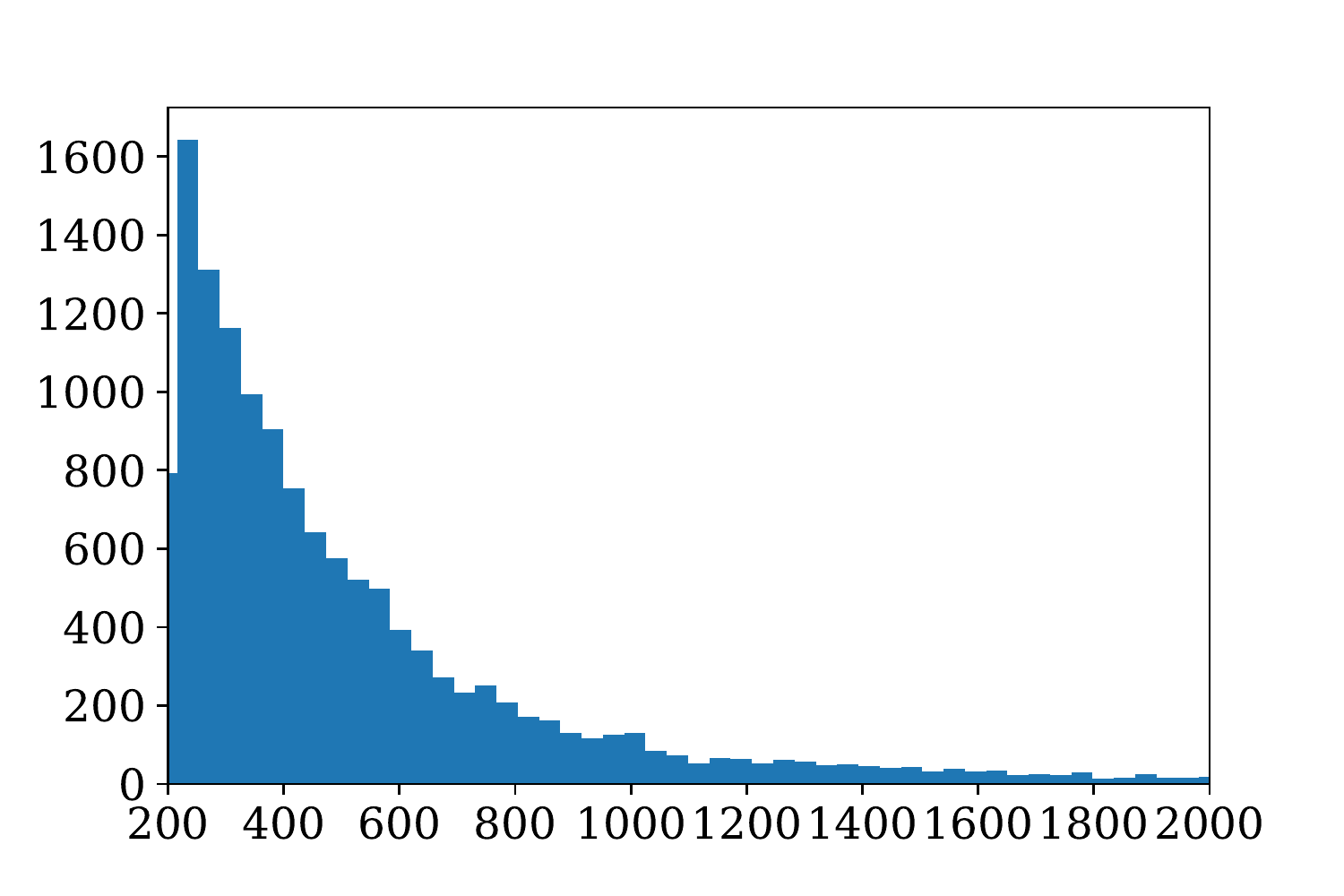}
Resolution

\textbf{1880 -- 2000}
\end{minipage}
\begin{minipage}{0.49\textwidth}
\centering
\includegraphics[width=\textwidth]{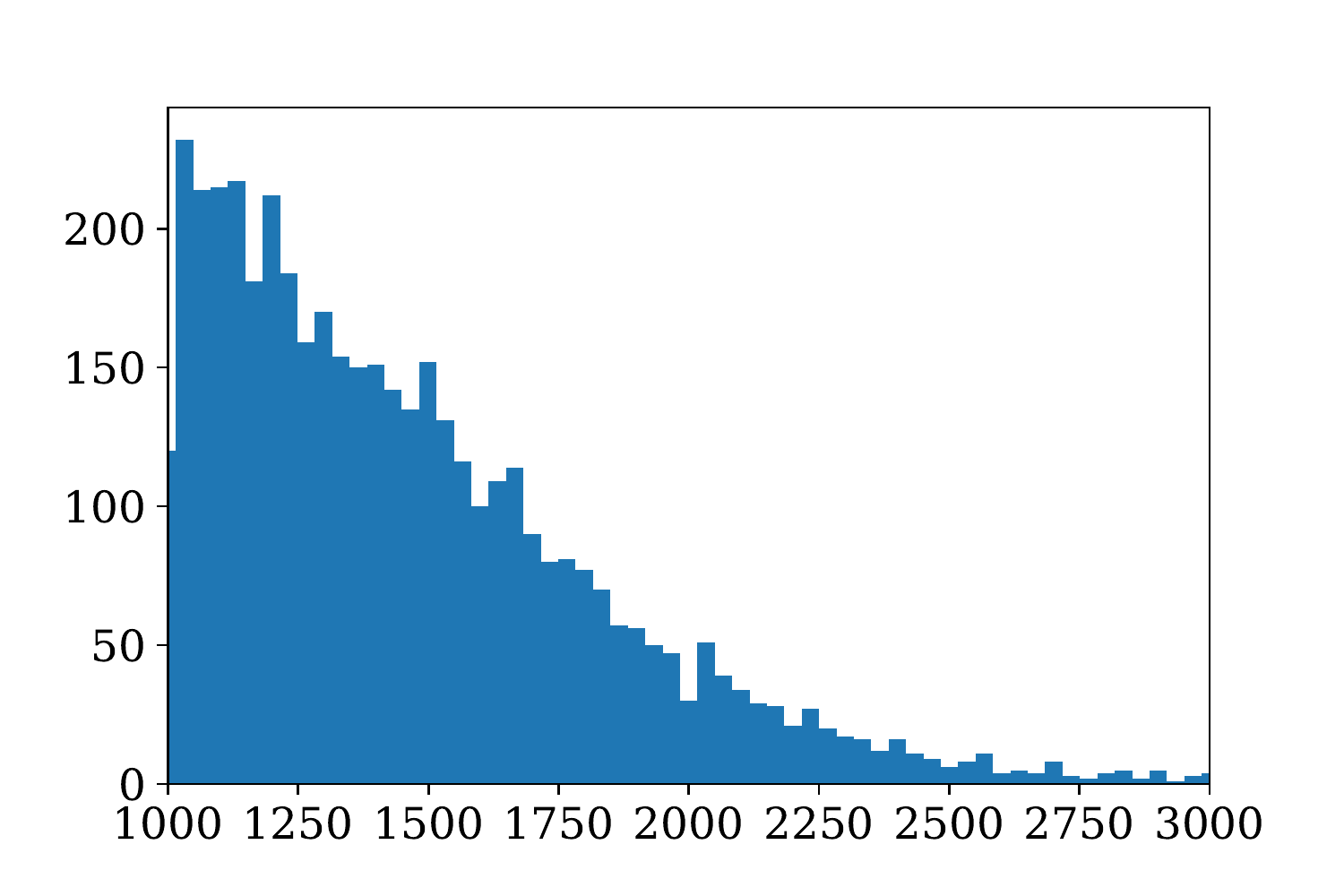}
Resolution

\textbf{2000 -- 2020}
\end{minipage}
\caption{\textbf{Resolution of Aligned Facial Images }. %
A histogram containing each image's resolution (measured in pixels) is illustrated above. For images from 1900 -- 2000, we only use images of resolution greater than or equal to $200\times200$.  Due to the abundance of digital images after 2000, we only use images of resolution greater than or equal to $1000\times1000$ in the 2000s and 2010s. Note that these resolutions correspond to the images after alignment and cropping is applied to extract normalized facial regions, and not to the original images found in Wikimedia Commons.}
  \label{fig:resolution} 
\end{figure}

\datasetname contains in total 26,247 portraits over 12 decades from the 1880s to 2010s. 
In Table \ref{tab:dataset_size}, we report the number of distinct identities per decade. Since our dataset contains only one image per identity, these numbers also correspond to the number of images per decade. 
We include a histogram over the image resolutions for all decades in Figure \ref{fig:resolution}.
From 1880--2000, our images have an average resolution of $527\times 527$. From 2000--2020, they have an average resolution of $1403\times1403$.

\begin{figure}[ht!]
\centering

\begin{minipage}{0.49\textwidth}
\centering
\includegraphics[width=\textwidth]{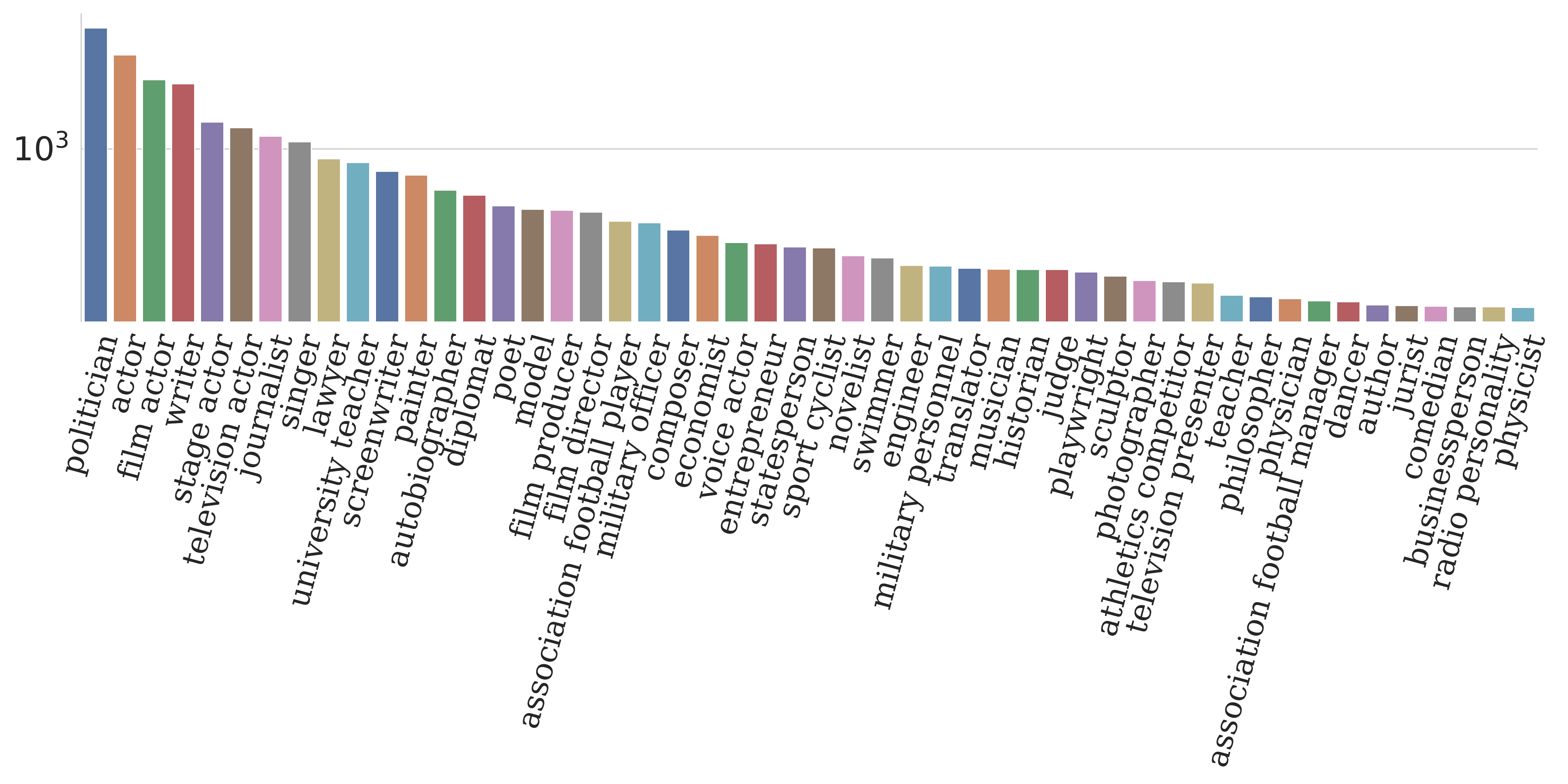}
\textbf{Occupations}
\end{minipage}
\begin{minipage}{0.49\textwidth}
\centering
\includegraphics[width=\textwidth]{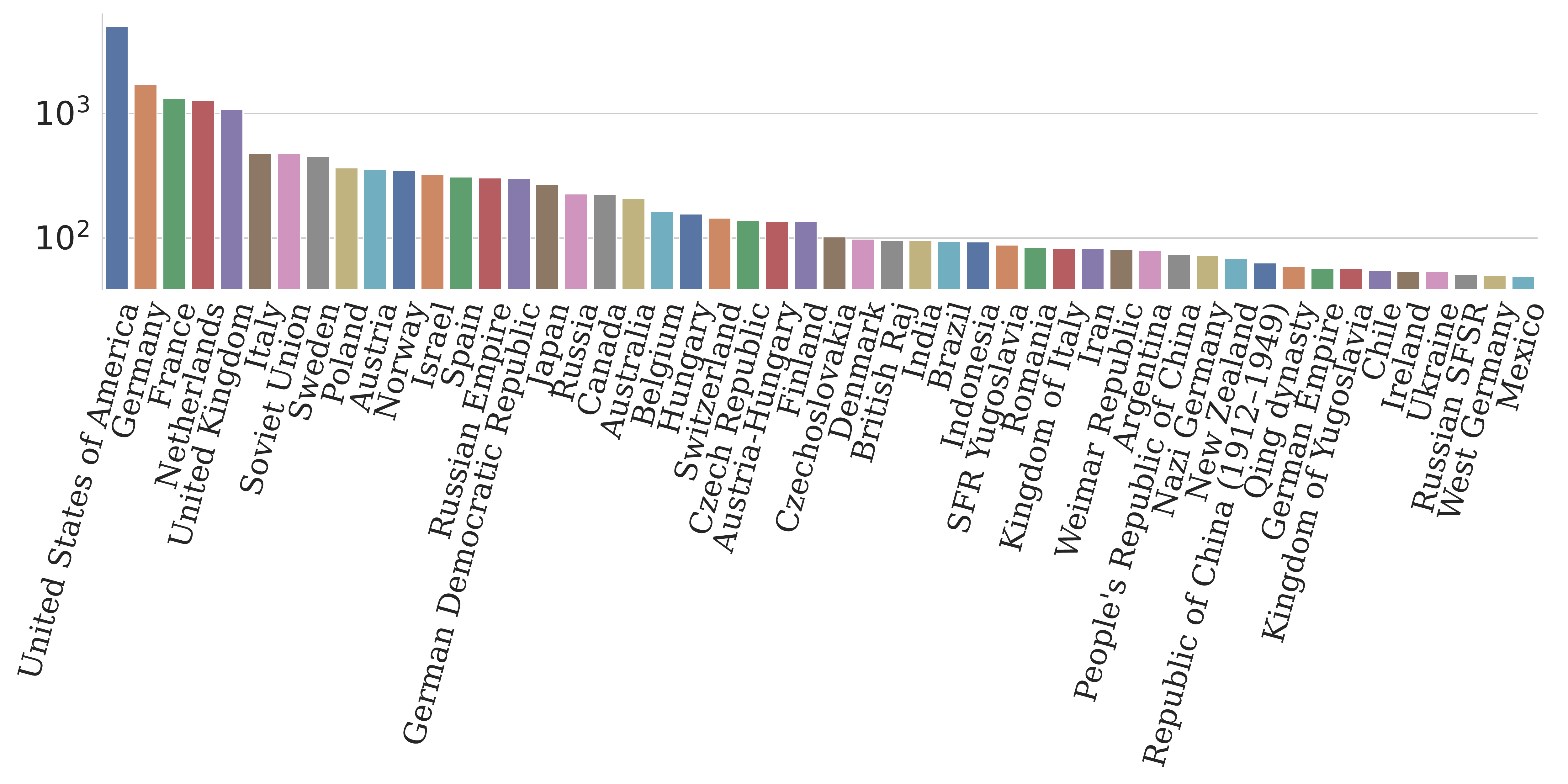}
\textbf{Citizenships}
\end{minipage}

\caption{\textbf{50 Most Common Occupations and Citizenships}. Identities may have more than one or no associated occupations and citizenships. There are 981 occupations not shown, each with 146 or fewer associated identities. Similarly, there are 209 citizenships not shown, each with 49 or fewer associated identities. Citizenship generally refer to historical nations. For example, individuals from what would be considered modern day China are categorized into the Qing dynasty, the Republic of China (1912-1949), the People's Republic of China, etc.
}
  \label{fig:citizenships}
\end{figure}

By associating the identity from each portrait with the biographic information on Wikimedia Commons, we show demographic information of \datasetname dataset in a set of histograms. Figure~\ref{fig:citizenships} shows the 50 most common citizenships and occupations. For a systematic review on the characteristics of the \datasetname dataset, please refer to the datasheet.

\newcolumntype{Y}{>{\centering\arraybackslash}X}
\newcolumntype{L}{>{\arraybackslash}X}
\begin{table*}[ht!]
\centering
\setlength{\tabcolsep}{2.8pt}
\def\arraystretch{1.0}
\begin{tabularx}{0.76\textwidth}{lcccccccccccccccc}
\toprule
    & &1880 & 1890 & 1900 & 1910 & 1920 & 1930 & 1940 & 1950 & 1960 & 1970 & 1980 & 1990 & 2000 & 2010   \\ \midrule
    Identities && 525 & 842 & 2049 & 3052 & 2042 & 1710 & 1337 & 1638 & 2611 & 1916 & 1382 & 928 & $3116^\star$ & $3099^\star$ \\
\bottomrule
\end{tabularx}
\caption{Number of identities per decade in the \datasetname{} dataset.
The symbol $^\star$ denotes that these sets were trimmed. Note that our dataset contains one image per identity, and therefore, these numbers also correspond to the number of images per decade.
}
\label{tab:dataset_size}
\end{table*}


\subsection{Assembly: Face Clustering}
We elaborate on our face clustering method (see Sec. 3 in the main paper) here. Although one could naively associate all faces in an identity's photos to that identity, a given photo may have multiple faces or may not picture the target identity at all. We, therefore, approach face assignment as a clustering problem, where we separately group faces corresponding to each person $i$. Let $F_i$ be the full set of faces extracted from images in $i$'s category. We wish to cluster $F_i$ into groups of people and identify the actual identity's group. We propose a hybrid semi-supervised clustering method for this problem, based on the following observations:
\begin{enumerate}
    \item The number of people represented in a set is unconstrained, so it is difficult to partition sets with parametric techniques, such as k-means, which assume a known number of clusters.
    \item A cluster is most likely to represent a single person if all face pairs are similar.
    \item If multiple faces appear in the same image, it is very unlikely that these faces belong to the same identity. As a corollary, given two images, each face in one of the images should only belong to the same identity as at most one face in the other image.
\end{enumerate}

\noindent \textbf{Maximal Clique Clustering.} We construct a graph $\mathcal{G}$ and formulate this as a graph clustering problem. Each face in $F_i$ corresponds to a node $p$ in $\mathcal{G}$. For each face, we also compute its FaceNet embedding $\mathbf{v}_p \in \mathbb{R}^{128}$ using OpenFace~\cite{amos2016openface}. By observation (2), our goal is then to find subsets of faces $C \subseteq \mathcal{G}$ such that every pair $p,q \in C$ is a positive verification pair ($\left\lVert\mathbf{v}_p - \mathbf{v}_q\right\rVert \leq \epsilon$ for some threshold $\epsilon$). Hence, we add an edge $(p,q)$ to $\mathcal{G}$ for such pairs.

However, we can further constrain this construction with observation (3). Let $I$ and $J$ be a pair of images, containing faces denoted by the sets $I_F$ and $J_F$. We observe that the corresponding subgraph of $\mathcal{G}$ induced by $I_F \cup J_F$ must be bipartite, and that at most one edge should have an endpoint in each node to represent correct identity relations. For each pair of images, we construct edges for each pair of nodes $(p \in I_F, q \in J_F)$ in increasing order by $d_{pq} = \left\lVert\mathbf{v}_p - \mathbf{v}_q\right\rVert$ if $d_{pq} \leq \epsilon$ and no edge adjacent to $p$ or $q$ already exists. Prior work shows that setting $\epsilon \approx 1$ is effective in practice~\cite{amos2016openface}, which we corroborated in our evaluation.

We would then like to search this graph for \emph{cliques}, subgraphs with an edge between every pair of nodes (specifically, maximal cliques, which cannot be further extended). We apply the Bron-Kerbosch~\cite{bron1973algorithm} algorithm to (relatively) efficiently enumerate all maximal cliques (this is an NP-complete problem~\cite{karp1972reducibility}). We sorted the resulting set of cliques in decreasing order by size, successively selecting those without nodes in any previously selected clique, until none remained. The selected cliques are our final clusters of $F_i$.

We further purified clusters by removing faces with an outlier threshold $\alpha$. Similar to the approach in MF2~\cite{nech2017level}, we first create a vector $v$ whose elements are the mean pairwise distance for each face in the cluster. We compute the median absolute deviation $\textrm{MAD}(v) = \textrm{Median}\left(|v - \textrm{Median}(v)|\right)$. Each face is an outlier if its corresponding $v_i$ satisfies $|v_i - \textrm{Median}(v)| \mathbin{/} \textrm{MAD}(v) > \alpha$ for $\alpha = 3.0$.

Since nearly all identities are annotated with a number of ground-truth reference images (i.e., their prominent Wikipedia article image or images listed in WikiData), we can simply assign the cluster containing the largest number of these reference images to that identity. If no references are available, we instead assume the largest cluster of faces is the person in question.
By visual inspection of 200 identities, we found over 98\% accuracy between clusters and their automatically assigned identity labels. We primarily attribute this to having reference images for over 86\% of identities.

\section{Method Details and Additional Results}\label{supp:results}
In this section, we include details and results that are omitted in the main paper due to the page limit. Note that these are supplementary to the main results, not to be viewed as significant new results.

\subsection{Detailed adaptation of state-of-the-art methods}

\medskip \noindent    \textbf{StarGAN v2~\cite{Choi2020StarGANVD}.} StarGAN is well-suited for our decade translation task because it scales to multiple domains. We ran the training code found in their \href{https://github.com/clovaai/stargan-v2}{official code repository}. We used the CelebA-HQ~\cite{CelebAMask-HQ} configuration. %

\begin{figure*}[t]
 \centering

\begin{minipage}{0.4\textwidth}
\jsubfig{\includegraphics[width=0.9\textwidth]{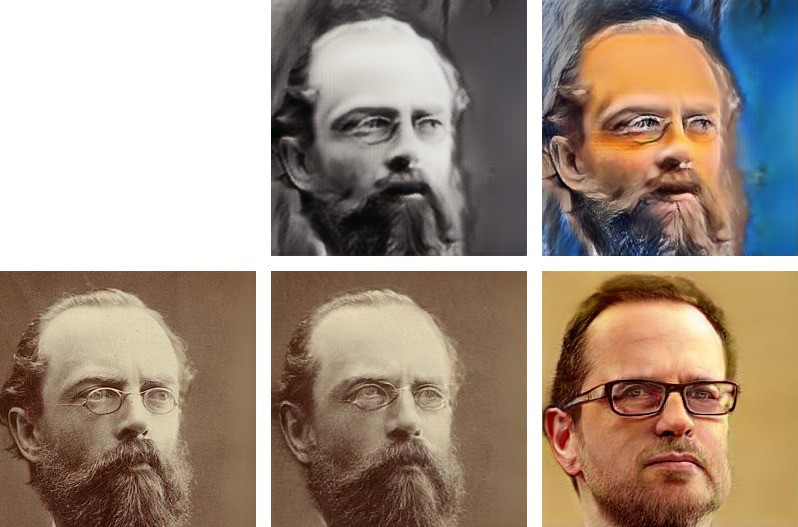}}{}
\begin{tabularx}{0.9\linewidth}{FFF}
Input (1900s) & 1900s & 2010s
\end{tabularx}
\end{minipage}
\begin{minipage}{0.4\textwidth}
\jsubfig{\includegraphics[width=0.9\textwidth]{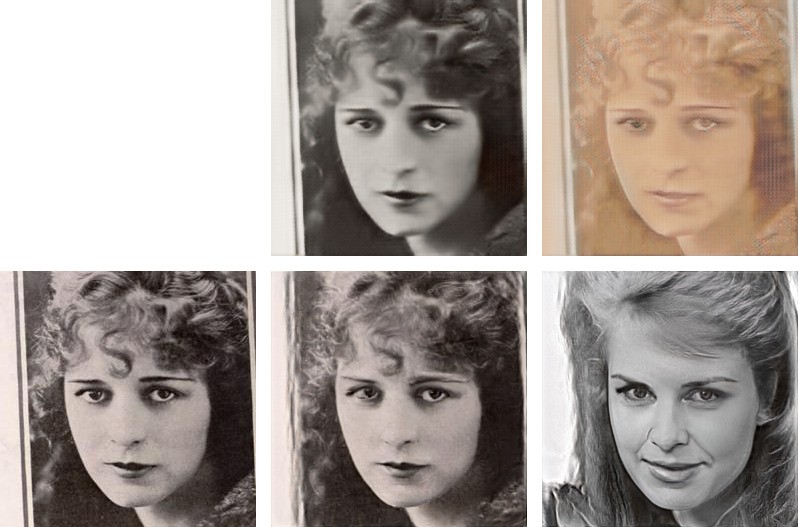}}{}
\rotatebox[origin=rB]{-90}{\begin{tabularx}{120px}{FFFF}DRIT++ & Ours \end{tabularx}}
\begin{tabularx}{0.9\linewidth}{FFF}
Input (1920s) & 1920s & 1960s
\end{tabularx}
\end{minipage}

\begin{minipage}{0.4\textwidth}
\jsubfig{\includegraphics[width=0.9\textwidth]{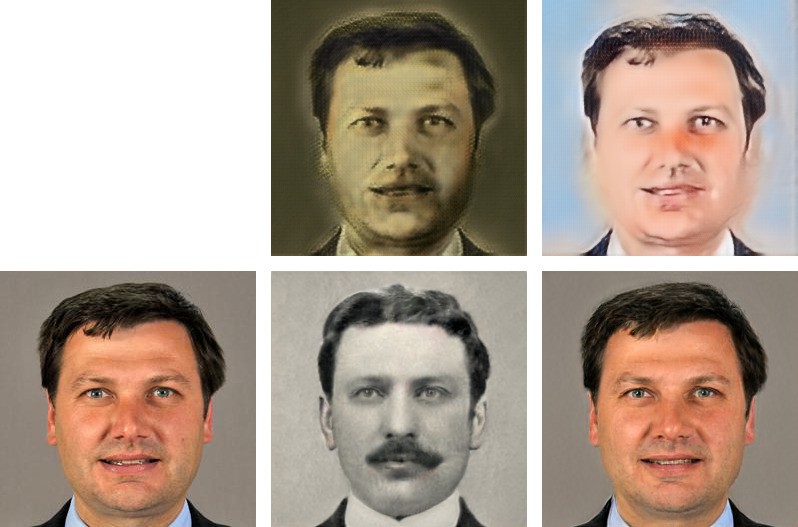}}{}
\begin{tabularx}{0.9\linewidth}{FFF}
Input (2010s) & 1900s & 2010s
\end{tabularx}
\end{minipage}
\begin{minipage}{0.4\textwidth}
\jsubfig{\includegraphics[width=0.9\textwidth]{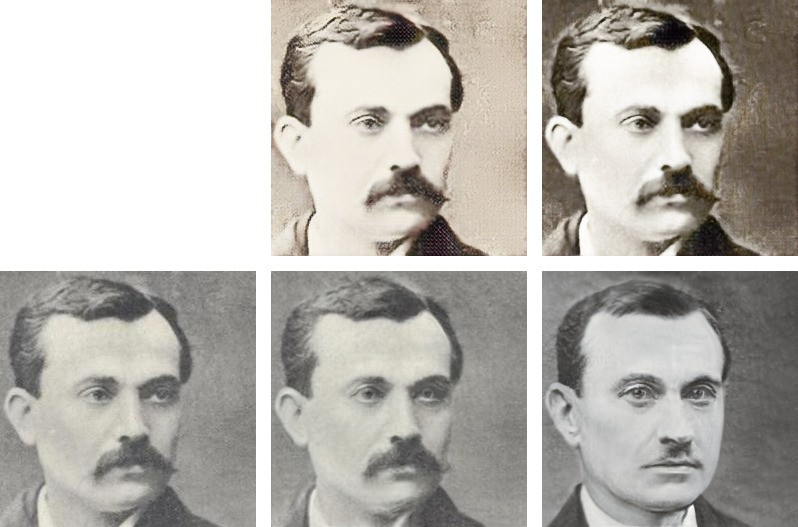}}{}
\rotatebox[origin=rB]{-90}{\begin{tabularx}{120px}{FFFF}DRIT++ & Ours \end{tabularx}}
\begin{tabularx}{0.9\linewidth}{FFF}
Input (1900s) & 1900s & 1920s
\end{tabularx}
\end{minipage}
    \caption{\textbf{DRIT++ Results.} We show additional qualitative results obtained using DRIT++ and our method. Compared to our method, DRIT has trouble reconstructing high quality images. In addition, most of the changes from DRIT are limited to color.}
    \label{fig:drit}
\end{figure*}

\medskip \noindent         \textbf{DRIT++~\cite{Lee2020DRITDI}.}
DRIT++ is a state-of-the-art image-to-image translation framework known for generating diverse representations. We train a model using the code in their \href{https://github.com/HsinYingLee/DRIT}{official code repository}. 
In practice, DRIT++ scales poorly to our problem because the model can only be trained on two domains at a time. 91 models are needed to create transformations across all 14 decades. To visualize results, we ran the model on several pairs of decades.
Evaluation shows that the model suffers from poor image quality compared to other baselines, as illustrated in Figure \ref{fig:drit}.

\medskip \noindent  \textbf{SAM~\cite{alaluf2021matter}.} SAM has shown impressive realism with regard to age transformation. We ran the training code found in their \href{https://github.com/yuval-alaluf/SAM}{official code repository}. We train SAM on top of a StyleGAN model trained on images from all decades in our dataset. During training, SAM searches for decade transformation directions within the model's $\mathcal{W}+$ space.  To modify SAM, we replace the age regression network with a decade classification network. %

 \medskip \noindent \textbf{StyleCLIP~\cite{Patashnik_2021_ICCV}.} Guided by CLIP~\cite{Radford2021LearningTV}, StyleCLIP  uses a text prompt to transform an image in StyleGAN's $\mathcal{W+}$ space. Similarly to SAM, we run StyleCLIP on a StyleGAN model trained on all of our dataset's images. Although StyleCLIP presents three different training schemes, we used the latent mapper approach since the authors claim that it is best suited for complex attributes. We ran the training code found in their \href{https://github.com/orpatashnik/StyleCLIP}{official code repository}. For evaluation, we set the target prompt as ``A person from the [decade]s" where decade is an element of $\{1880, 1890, \dots, 2010\}$. %
  
\medskip \noindent \textbf{StyleGAN-nada~\cite{gal2021stylegannada}.} 
Because StyleGAN-nada is designed for out of domain changes, we started with images from FFHQ and projected them to decades in the 20th century. We used the same text prompt that we used for StyleCLIP. We ran the training code found in their \href{https://github.com/rinongal/StyleGAN-nada}{official code repository}. We also experimented with using exemplar images, and found that the generated images suffer from a lack of style diversity and are entangled with the identity of the exemplars, meaning that synthesized images adopted facial characteristics of the exemplar images.

\subsection{Decade-classification performance}
\begin{figure}[t]
    \centering
    \includegraphics[width=0.5\textwidth]{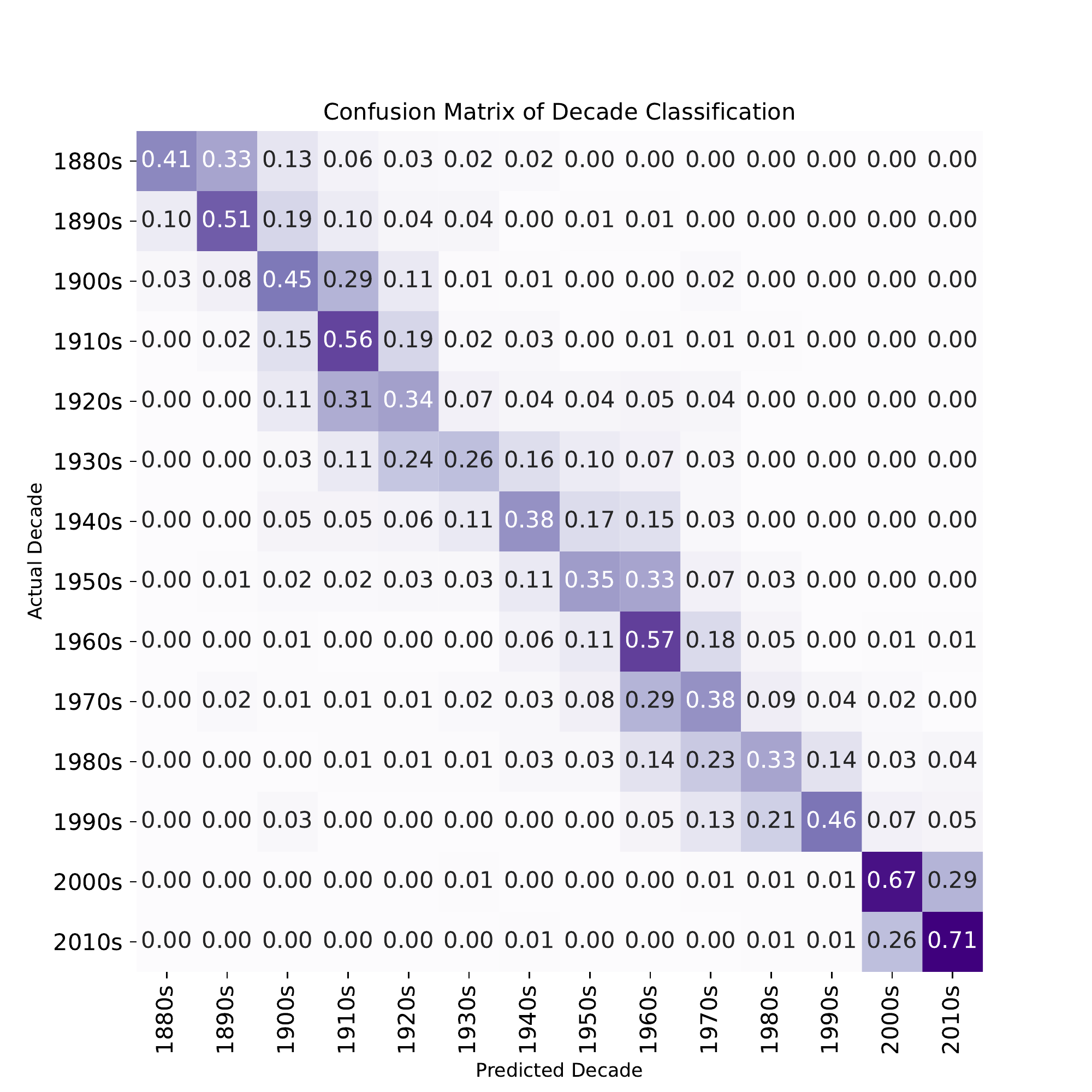}
    \caption{Test classification accuracy on the \datasetname{} test set. Rows indicate the ground truth decade and columns indicate the predicted decade.}
    \label{fig:classifier-confusion}
\end{figure}

In our evaluation, we use an EfficientNetB0 classification network trained on the \datasetname dataset to calculate the DCA scores. 
For reference, a confusion matrix on the test set of this classifier is in Figure \ref{fig:classifier-confusion}. 
The classifier has an average accuracy of $45.57\%$ over all decades. Furthermore, $79.21\%$ of the confusion is captured within a tolerance of $\pm 1$ decade. 

\subsection{Additional comparisons}
Comparisons between our model and state-of-the-art alternatives on the full test set of \datasetname are provided separately using an interactive viewer. 
Consistent with the results presented in the main paper, our method outperforms alternatives in terms of image quality and style changes, while preserving the identity of the input images. 

\begin{figure*}[ht!]
\centering
\includegraphics[width=0.49\linewidth]{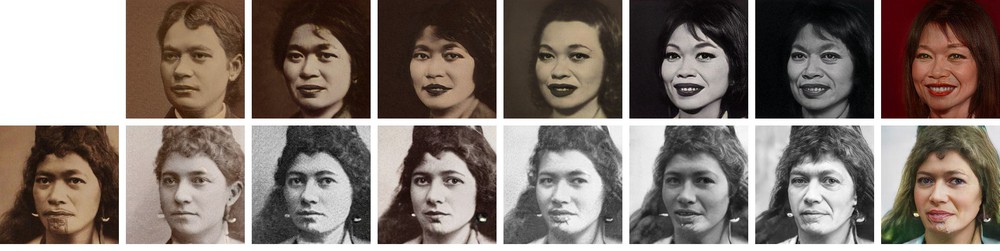}
\includegraphics[width=0.49\linewidth]{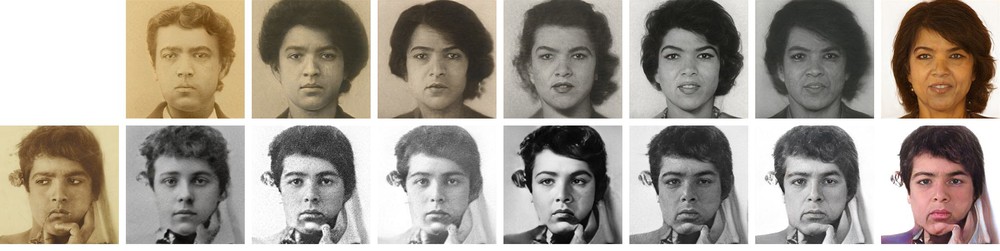}
\vspace{5pt}

\includegraphics[width=0.49\linewidth]{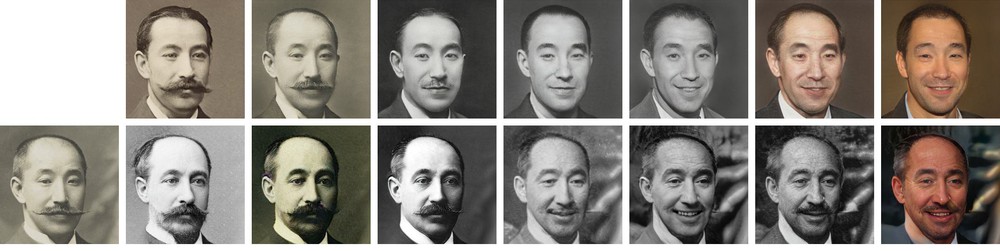}
\includegraphics[width=0.49\linewidth]{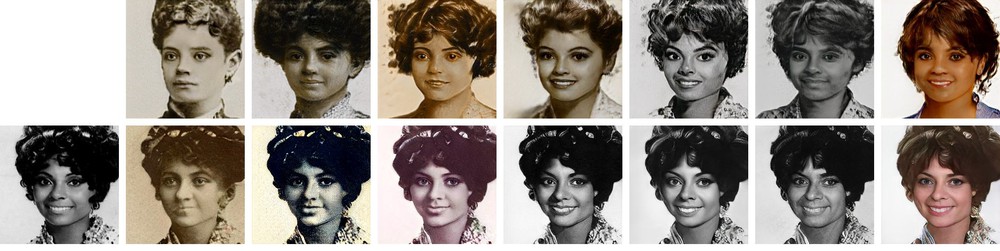}
\vspace{5pt}

\includegraphics[width=0.49\linewidth]{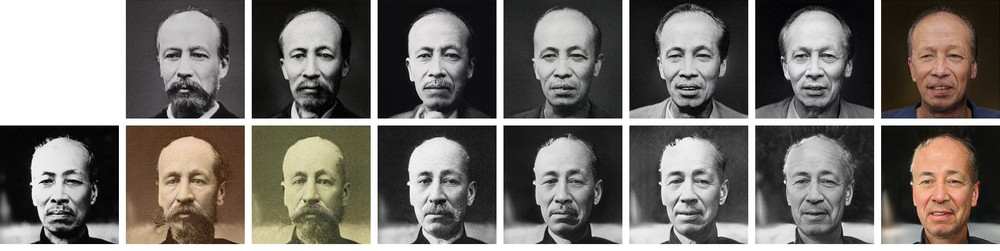}
\includegraphics[width=0.49\linewidth]{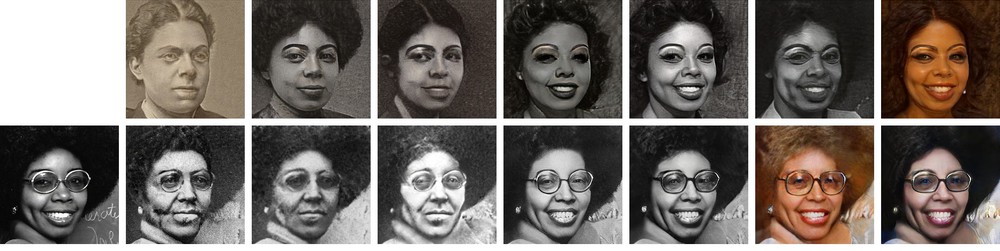}
\vspace{5pt}

\includegraphics[width=0.49\linewidth]{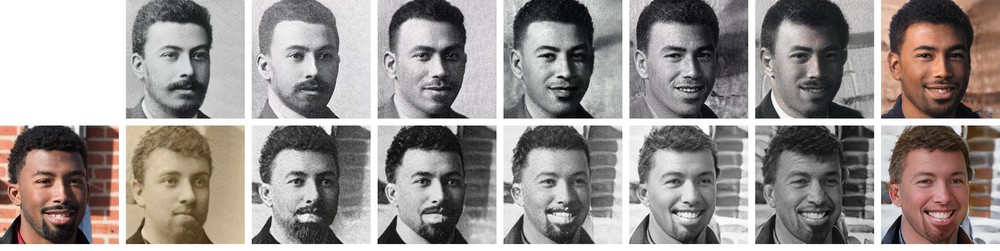}
\includegraphics[width=0.49\linewidth]{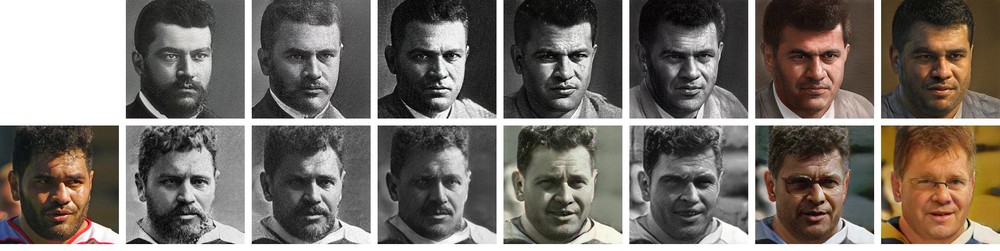}
\vspace{5pt}

\includegraphics[width=0.49\linewidth]{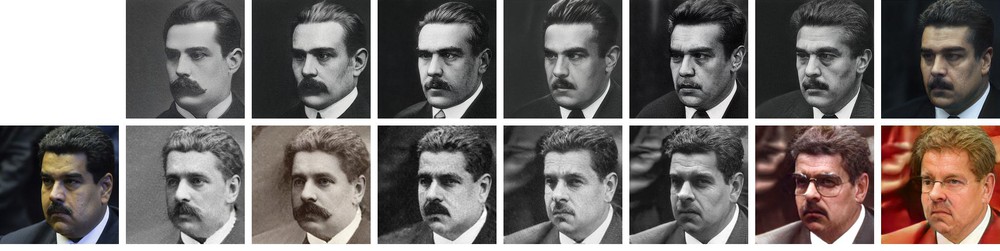}
\includegraphics[width=0.49\linewidth]{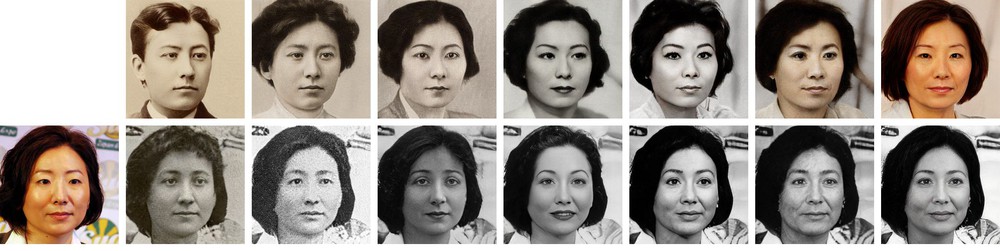}

\begin{tabularx}{0.49\linewidth}{FFFFFFFF}
Input & 1880 & 1900 & 1920 & 1940 & 1960 & 1980 & 2000
\end{tabularx}
\begin{tabularx}{0.49\linewidth}{FFFFFFFF}
Input & 1880 & 1900 & 1920 & 1940 & 1960 & 1980 & 2000
\end{tabularx}

\caption{\textbf{Comparison with StarGAN}. We highlight differences between our method (first row) and StarGAN~\cite{Choi2020StarGANVD} (second row) on a selection of individuals balanced by gender and ethnicity. We show that our method accentuates differences in style. While StarGAN is able to generate plausible transformations, it has a poor understanding of skin tone and overall identity, which is especially critical for real-world applications. } 
  \label{fig:stargan-comparison}
\end{figure*}

\begin{figure*}[h]
\centering
\jsubfig{\includegraphics[width=\linewidth]{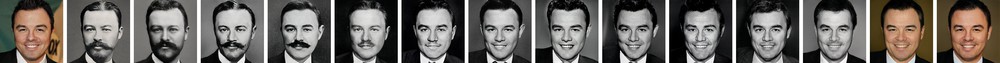}}{}
\newline

\jsubfig{\includegraphics[width=\linewidth]{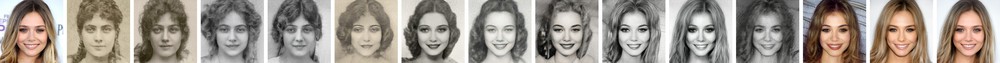}}{}
\newline

\jsubfig{\includegraphics[width=\linewidth]{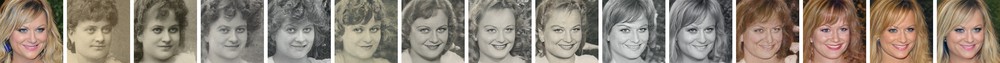}}{}
\newline

\jsubfig{\includegraphics[width=\linewidth]{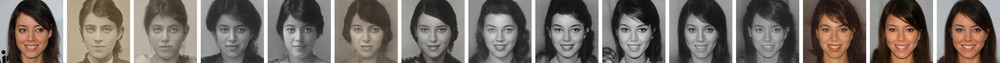}}{}
\newline

\jsubfig{\includegraphics[width=\linewidth]{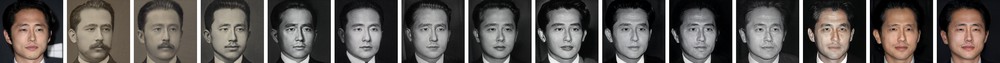}}{}
\newline

\jsubfig{\includegraphics[width=\linewidth]{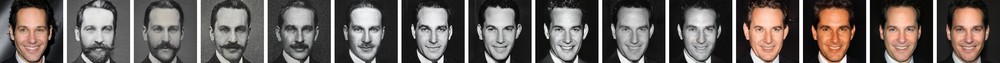}}{}
\newline

\jsubfig{\includegraphics[width=\linewidth]{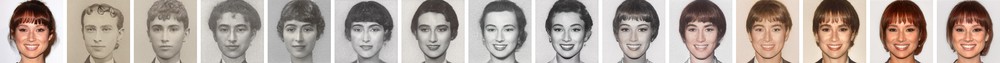}}{}
\newline

\jsubfig{\includegraphics[width=\linewidth]{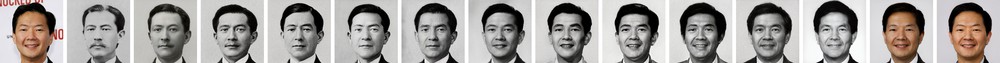}}{}
\newline

\jsubfig{\includegraphics[width=\linewidth]{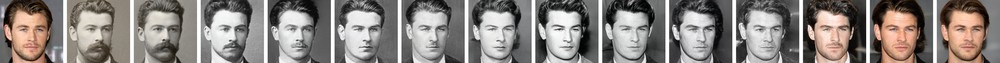}}{}
\newline

\jsubfig{\includegraphics[width=\linewidth]{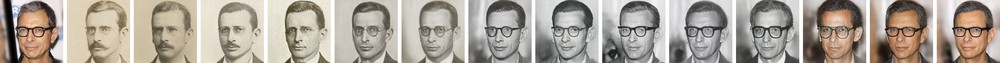}}{}
\newline

\jsubfig{\includegraphics[width=\linewidth]{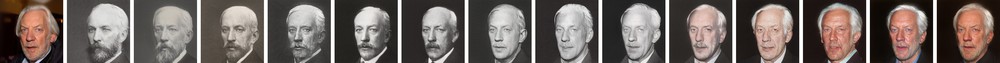}}{}
\newline

\jsubfig{\includegraphics[width=\linewidth]{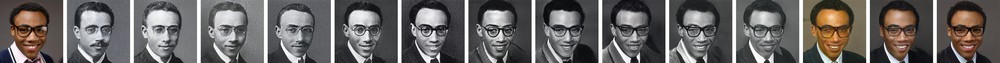}}{}

\begin{tabularx}{\linewidth}{FFFFFFFFFFFFFFF}
Input & 1880s & 1890s & 1900s & 1910s & 1920s & 1930s & 1940s & 1950s& 1960s & 1970s & 1980s  & 1990s & 2000s & 2010s
\end{tabularx} 
    \caption{\textbf{Results on CelebAHQ}. Above we show results on celebrities from CelebAHQ. We see that the celebrities remain recognizable throughout the transformations.}
    \label{fig:celebahq}
\end{figure*}

\begin{figure*}[ht!]
\centering

\jsubfig{
\includegraphics[width=0.25\linewidth]{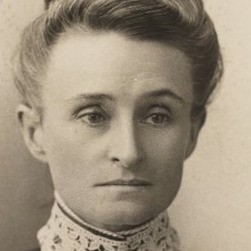}
\includegraphics[width=0.25\linewidth]{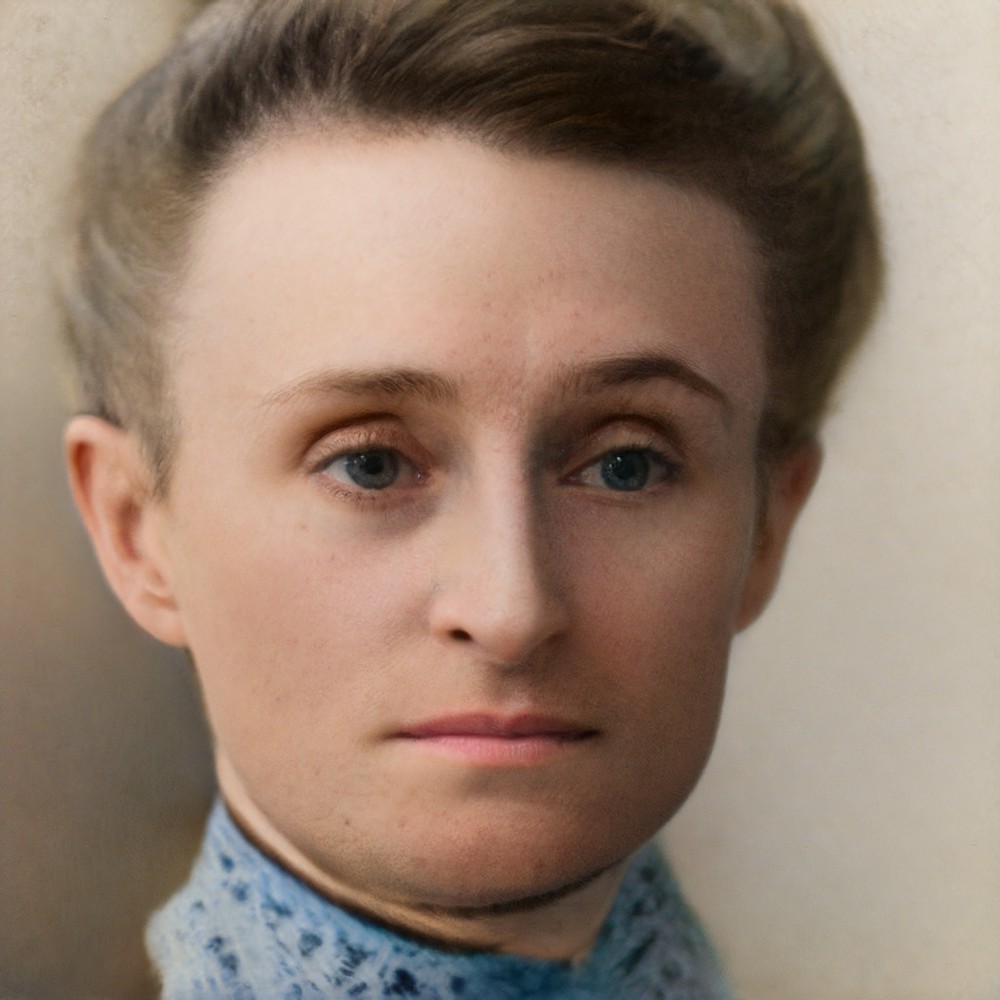}
\includegraphics[width=0.25\linewidth]{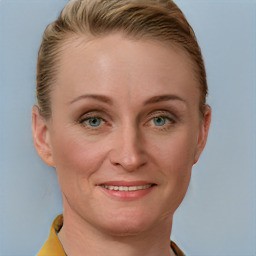}
}{}
\jsubfig{
\includegraphics[width=0.25\linewidth]{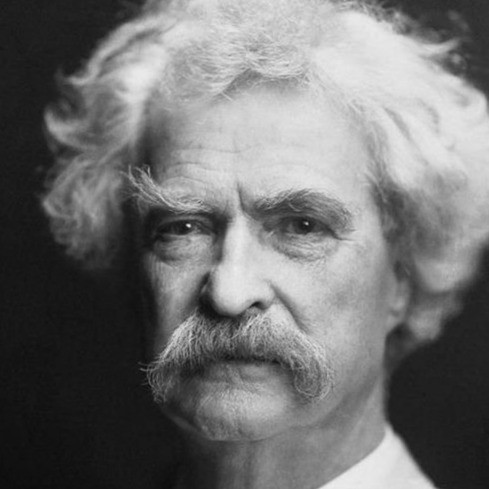}
\includegraphics[width=0.25\linewidth]{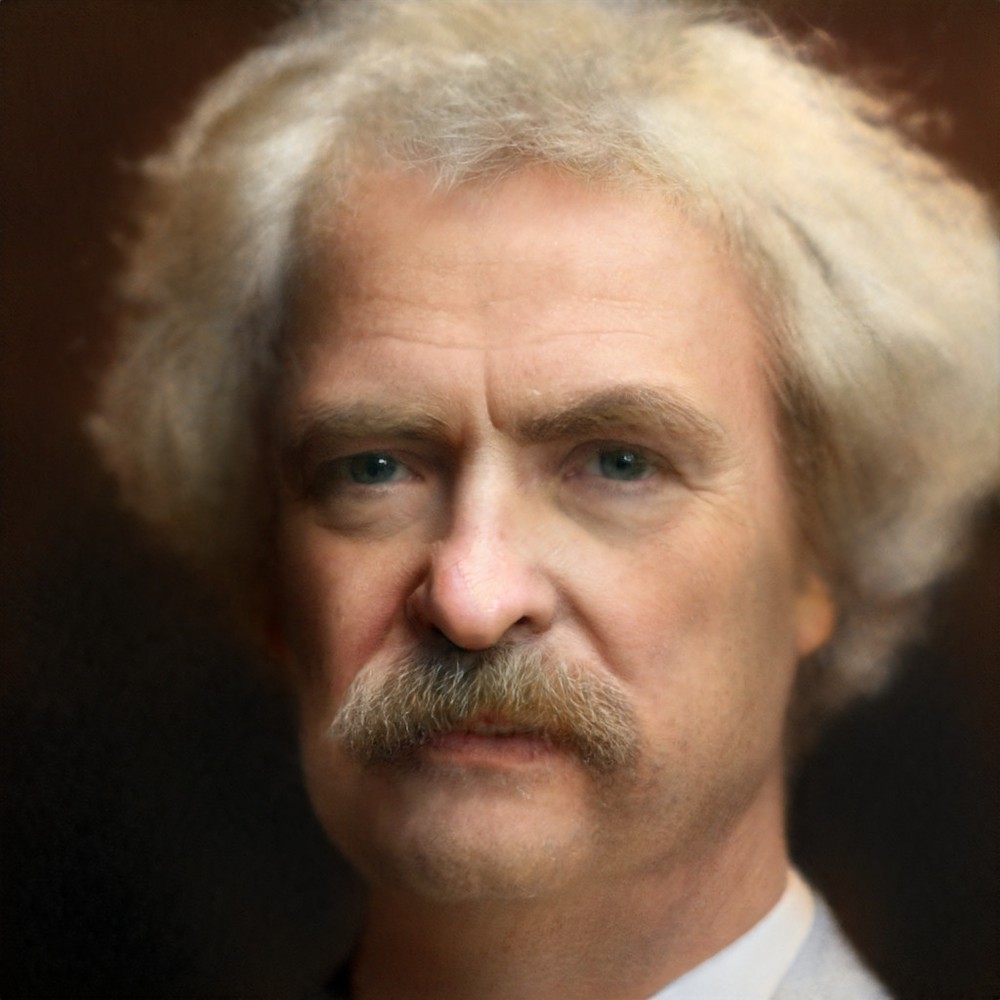}
\includegraphics[width=0.25\linewidth]{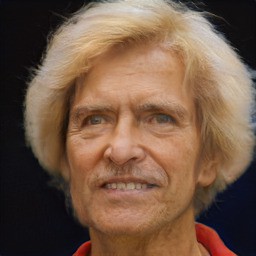}
}{}
\jsubfig{
\includegraphics[width=0.25\linewidth]{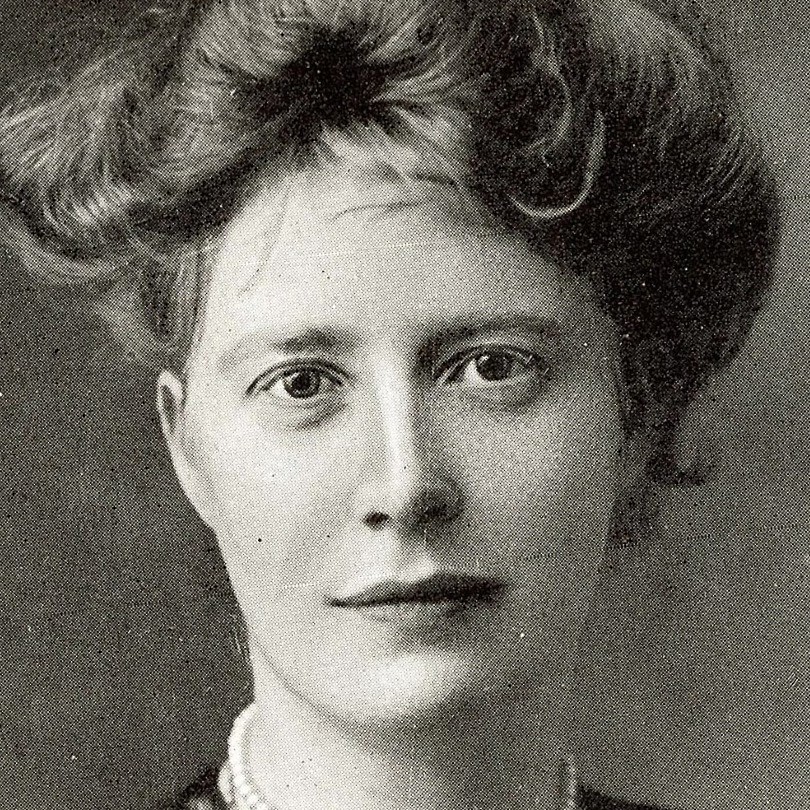}
\includegraphics[width=0.25\linewidth]{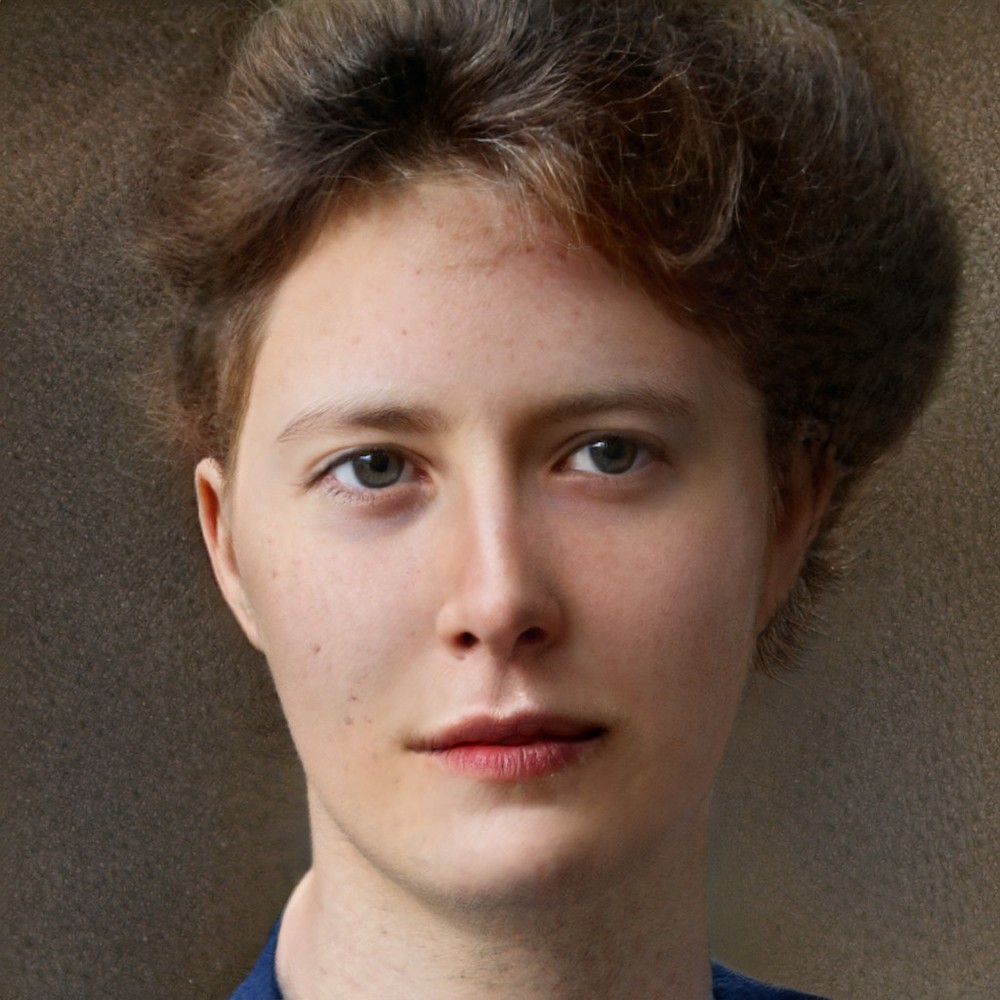}
\includegraphics[width=0.25\linewidth]{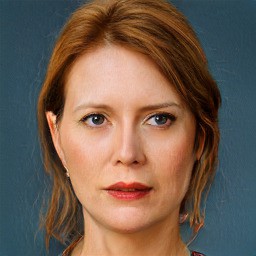}
}{}
\begin{tabularx}{0.75\linewidth}{FFF}
Input & Time Travel Rephotography \cite{Luo-Rephotography-2021} & Ours
\end{tabularx}

\caption{\textbf{Comparison with Time Travel Rephotography}. We show our method on images taken from the supplementary material of Time Travel Rephotography \cite{Luo-Rephotography-2021}. For our method, we transform each face to the 2010s. As illustrated above, our method modifies the style to simulate what these individuals would have looked like had they lived today (and also allow for other transformations through time), whereas \cite{Luo-Rephotography-2021} focuses primarily on image restoration.  }
  \label{fig:rephotography}
\end{figure*}

\begin{figure*}[h]
\centering
\begin{minipage}[]{0.8\linewidth}
\centering{\textit{Yearbook}~\cite{ginosar2015century}

\includegraphics[width=\linewidth]{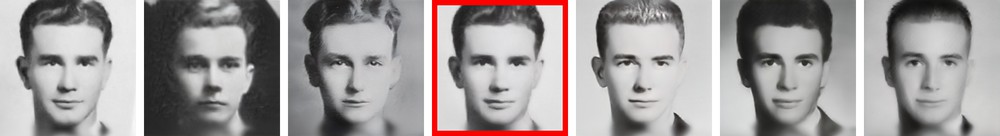}
\includegraphics[width=\linewidth]{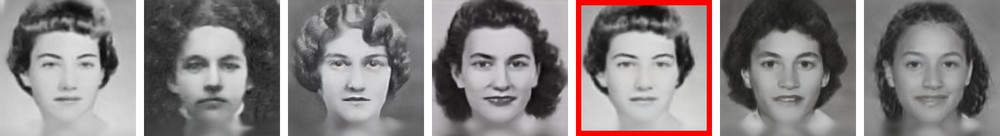}
\includegraphics[width=\linewidth]{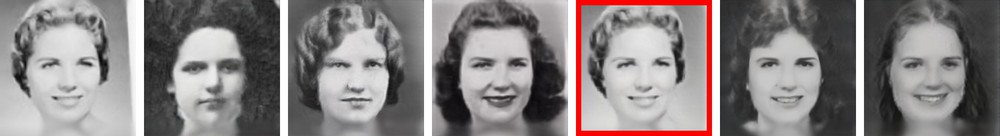}

\centering{\datasetname}

\includegraphics[width=\linewidth]{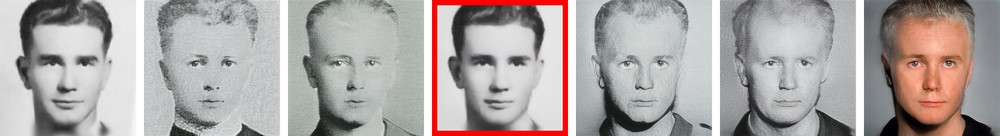}
\includegraphics[width=\linewidth]{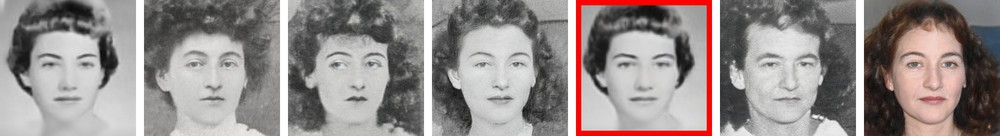}
\includegraphics[width=\linewidth]{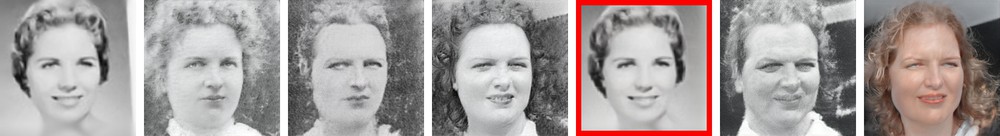}
\begin{tabularx}{\linewidth}{FFFFFFF}
Input & 1900s & 1920s & 1940s& 1960s& 1980s& 2000s
\end{tabularx}
}
\end{minipage}

\caption{\textbf{Results on the Yearbook dataset}. Given the input images from the left, which are sourced from the Yearbook dataset \cite{ginosar2015century}, we compare the quality of transformations between a model trained on the Yearbook dataset and models trained on \datasetname. Because the Yearbook dataset photos are lower resolution, we decided to not include them in the \datasetname dataset.}
  \label{fig:yearbook}
\end{figure*}

 We highlight differences between our model and StarGAN~\cite{Choi2020StarGANVD} in Figure~\ref{fig:stargan-comparison}. 
We present results on CelebAHQ~\cite{CelebAMask-HQ}, a dataset of recognizable celebrities, in Figure~\ref{fig:celebahq}.We also show comparisons with Time Travel Rephotography in Figure~\ref{fig:rephotography}, and the issues we experience with quality using Yearbook dataset in Figure~\ref{fig:yearbook}.

\subsection{$\mathcal{W}$ space vs. $\mathcal{W+}$ space}
\begin{figure*}[ht!]
\centering{
\vspace{-4pt}
$\mathcal{W}$ Inversion

\vspace{5px}
\jsubfig{\includegraphics[width=0.75\linewidth]{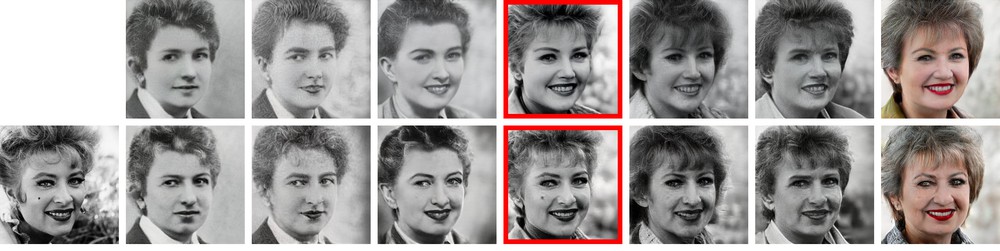}}{}
\rotatebox[origin=rB]{-90}{\begin{tabularx}{100px}{FF} Before \methodshort{} & After \methodshort{}\end{tabularx}}

$\mathcal{W+}$ Inversion

\vspace{5px}
\jsubfig{\includegraphics[width=0.75\linewidth]{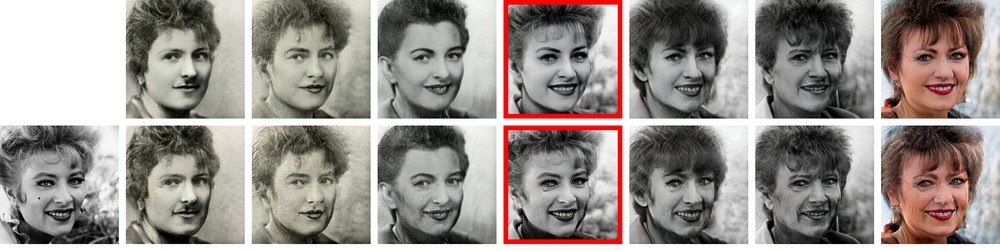}}{}
\rotatebox[origin=rB]{-90}{\begin{tabularx}{100px}{FF} Before \methodshort{} & After \methodshort{}\end{tabularx}}

$\mathcal{W}$ Inversion

\vspace{5px}
\jsubfig{\includegraphics[width=0.75\linewidth]{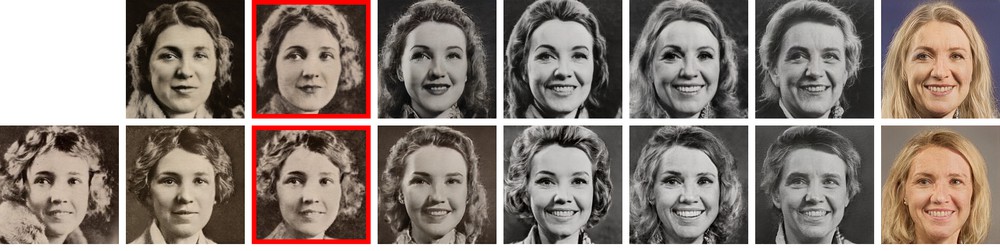}}{}
\rotatebox[origin=rB]{-90}{\begin{tabularx}{100px}{FF} Before \methodshort{} & After \methodshort{}\end{tabularx}}

$\mathcal{W+}$ Inversion

\vspace{5px}
\jsubfig{\includegraphics[width=0.75\linewidth]{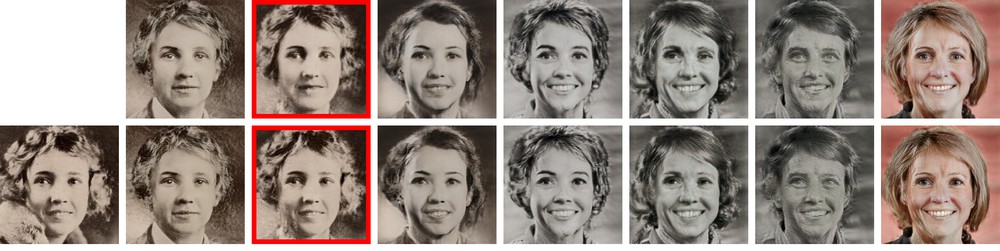}}{\begin{tabularx}{\linewidth}{FFFFFFFF}
Input & 1910s & 1920s & 1940s & 1960s & 1970s & 1980s & 2010s 
\end{tabularx}}
\rotatebox[origin=rB]{-90}{\begin{tabularx}{100px}{FF} Before \methodshort{} & After \methodshort{}\end{tabularx}}

}

\caption{\textbf{Face inversion and transformation results using the $\mathcal{W}$ vs. $\mathcal{W+}$ space}. Above we compare before \methodshort{} and after \methodshort{} results obtained using the $\mathcal{W}$ space, which we adopt in our work, with results obtained using the $\mathcal{W+}$ space. As demonstrated above, results with the $\mathcal{W+}$ space yield various artifacts, which are often amplified after \methodshort{}.  }
  \label{fig:w_plus}
\end{figure*}

As mentioned in the main paper, our method uses a $\mathcal{W}$ projection to invert an image into the latent space of a StyleGAN model. 
Figure \ref{fig:w_plus} shows a comparison to the alternative $\mathcal{W+}$ space. 
We find that inverting images into the $\mathcal{W+}$ space creates more artifacts during training, which are often amplified after \methodshort{}.  

\subsection{Additional ablation results}
\begin{figure*}[tb]
\jsubfig{\includegraphics[width=0.45\linewidth]{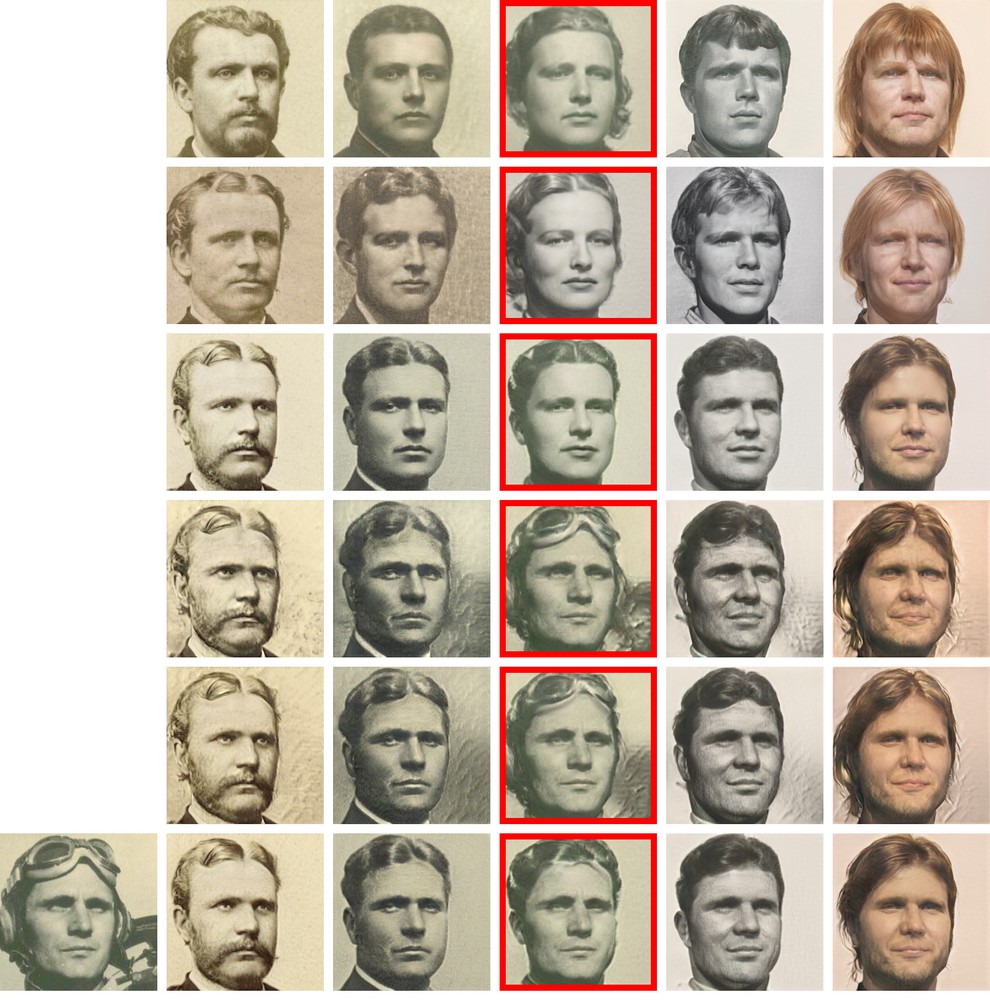}}{}
\hfill
\jsubfig{\includegraphics[width=0.45\linewidth]{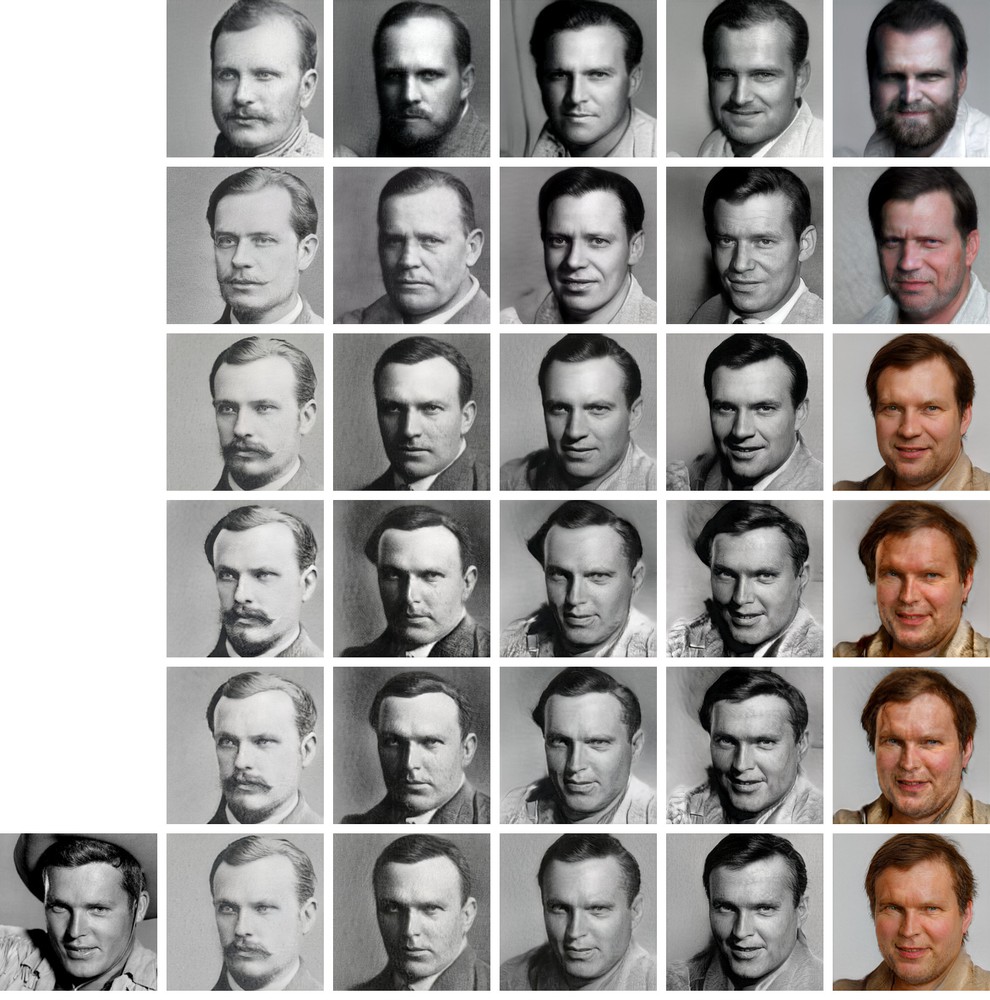}}{}
\rotatebox[origin=rB]{-90}{\begin{tabularx}{230px}{FFFFFF}No $\mathcal{L}_{\text{id}}^{\text{(I)}}$ &No LS & No TMT & No $\mathcal{L}_{\text{id}}^{\text{(II)}}$ & No Mask& Ours \end{tabularx}}

\jsubfig{\includegraphics[width=0.45\linewidth]{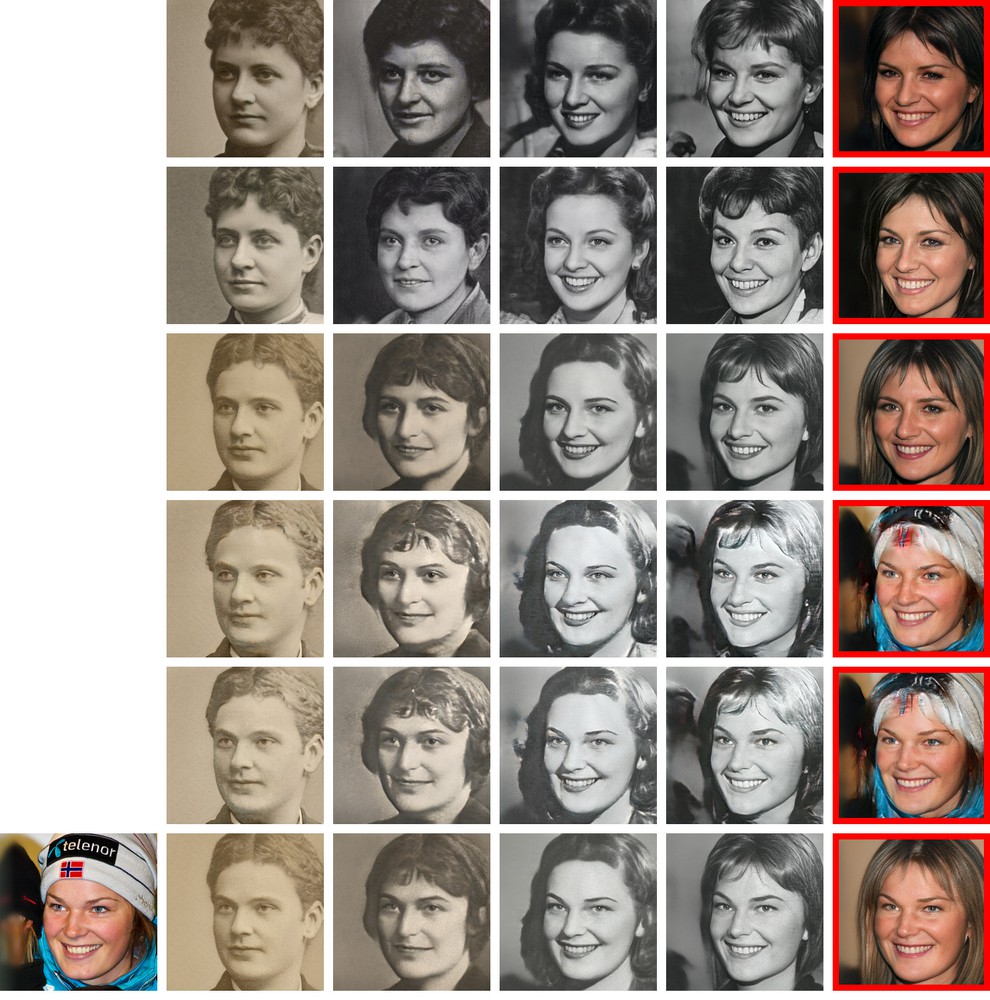}}{}
\hfill\jsubfig{\includegraphics[width=0.45\linewidth]{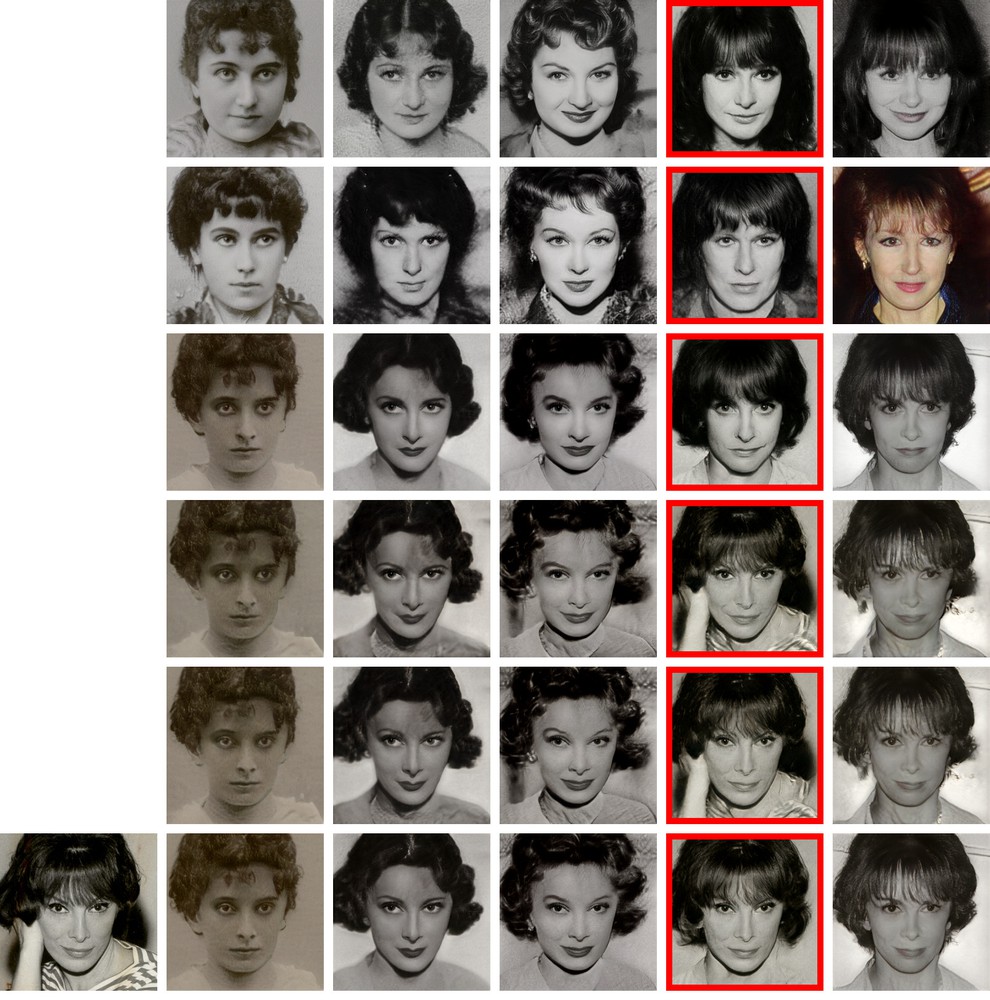}}{}
\rotatebox[origin=rB]{-90}{\begin{tabularx}{230px}{FFFFFF}No $\mathcal{L}_{\text{id}}^{\text{(I)}}$ &No LS & No TMT & No $\mathcal{L}_{\text{id}}^{\text{(II)}}$ & No Mask& Ours \end{tabularx}}
\begin{tabularx}{0.45\linewidth}{FFFFFF}
Input & 1880s & 1920s & 1940s & 1960s & 2000s
\end{tabularx}\hfill
\begin{tabularx}{0.45\linewidth}{FFFFFF}
Input & 1890s & 1910s & 1930s & 1950s & 1970s
\end{tabularx}
\caption{\textbf{Additional Ablation Results}. These results further show the improvement obtained in terms of identity preservation after adding $\mathcal{L}_{\text{id}}^{\text{(I)}}$, layer swapping, and \methodshort{} during training; and the benefit of incorporating $\mathcal{L}_{\text{id}}^{\text{(II)}}$ and the masking procedure. }
  \label{fig:appendix_ablations}
\end{figure*}

Figure \ref{fig:appendix_ablations} shows additional examples of ablations on essential components in our approach.
Consistent with the results in the main paper, our full model with all components enabled has the best performance compared to other variants.

\subsection{Analysis of \methodshort{} offsets}
Figure \ref{fig:pti_appendix} shows the effects of applying the proposed \methodshort{} offsets to portraits.
In general, we find that the \methodshort{} offsets are distinguishable from directions between arbitrary pairs of decade generators. This agrees with our intuition that offsets learned by fine-tuning a generator on an identity should be independent from the style changes between decades.

\section{Training and implementation details}\label{supp:details}

\textbf{Learning Decade Models.} All models were trained for 645k iterations on a single Nvidia RTX 3090. The codebase is derived from the official \href{https://github.com/NVlabs/stylegan2-ada-pytorch}{stylegan2-ada-pytorch repository}. For training, we use the $\texttt{paper256}$ config. We use the PyTorch implementation of \href{https://github.com/deepinsight/insightface}{InsightFace} with a ResNet-100 backbone for the identity loss. The identity loss was added to the $G_\texttt{main}$ phase of StyleGAN training with a weight of $1.0$. We used a regularization weight of $\gamma=0.5$, which we empirically found was best for the dataset. 

\medskip
\noindent \textbf{Single-image Refinement.} We modified the Pivotal Tuning Inversion (PTI) \href{https://github.com/danielroich/PTI}{code base} for our task. We added a DeepLab~\cite{deeplabv3plus2018} mask to the input images during the generator tuning phase of PTI. From experimentation, we found that the L2 loss had little effect on the quality of images. Most of the inversion tuning is guided by the LPIPS~\cite{zhang2018perceptual} loss. Because of this, we set a small LPIPS threshold of $0.03$ during training. Inference takes 2-3 minutes per image.

\newcolumntype{s}{>{\hsize=.8\hsize}X}
\newcolumntype{d}{>{\hsize=1.2\hsize}X}
\begin{figure*}[ht!]
    \centering
    \begin{tabularx}{\textwidth}{FF}
\includegraphics[width=0.35\linewidth]{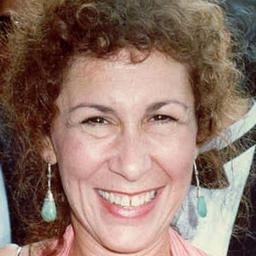}
    &\includegraphics[width=0.35\linewidth]{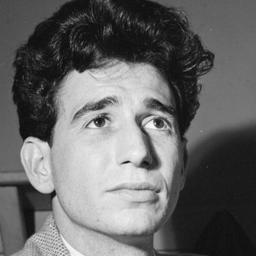}\\Input (1980s) &  Input (1940s)
    \end{tabularx}
    
    \begin{tabularx}{\textwidth}{FFFF}
    
    \includegraphics[width=0.7\linewidth]{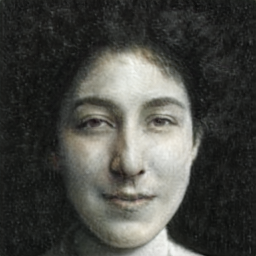} &
    \includegraphics[width=0.7\linewidth]{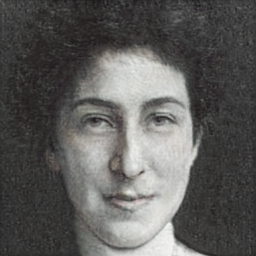} & 
    \includegraphics[width=0.7\linewidth]{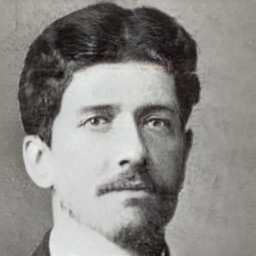} &
    \includegraphics[width=0.7\linewidth]{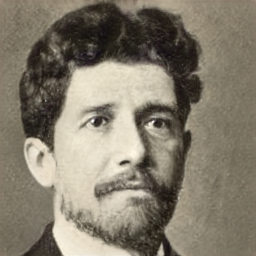}\\
    $ G_{1900}(w_1; \theta_{1900})$  &$ G_{1900}(w_1; \theta_{1900} + \Delta \theta_1)$ & $ G_{1900}(w_2; \theta_{1900})$  &$ G_{1900}(w_2; \theta_{1900} + \Delta \theta_2)$\\
    \includegraphics[width=0.7\linewidth]{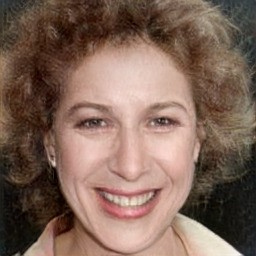} &
    \includegraphics[width=0.7\linewidth]{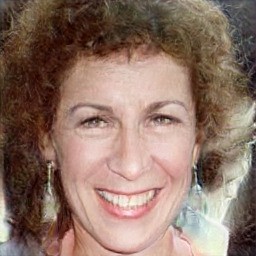} & 
    \includegraphics[width=0.7\linewidth]{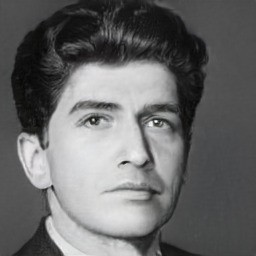} &
    \includegraphics[width=0.7\linewidth]{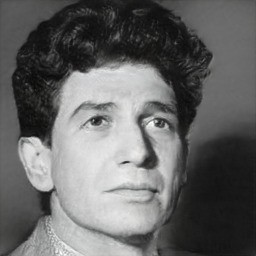}\\
    $ G_{1980}(w_1; \theta_{1980})$  &$ G_{1980}(w_1; \theta_{1980} + \Delta \theta_1)$ & $ G_{1940}(w_2; \theta_{1940})$  &$ G_{1940}(w_2; \theta_{1940} + \Delta \theta_2)$\\
     
    \includegraphics[width=0.7\linewidth]{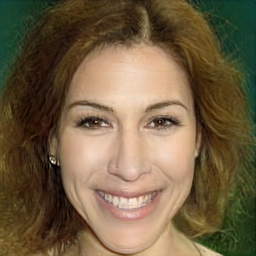} &
    \includegraphics[width=0.7\linewidth]{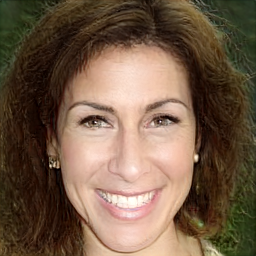} & 
    \includegraphics[width=0.7\linewidth]{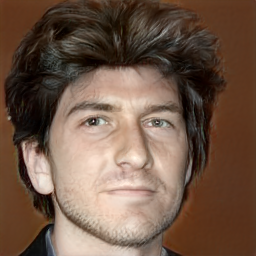} &
    \includegraphics[width=0.7\linewidth]{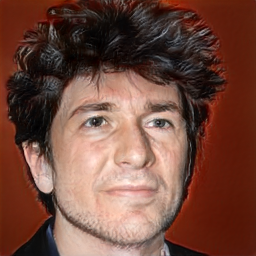}\\
    $ G_{2000}(w_1; \theta_{2000})$  &$ G_{2000}(w_1; \theta_{2000} + \Delta \theta_1)$ & $ G_{2000}(w_2; \theta_{2000})$  &$ G_{2000}(w_2; \theta_{2000} + \Delta \theta_2)$
    \end{tabularx}
    \includegraphics[width=0.49\textwidth]{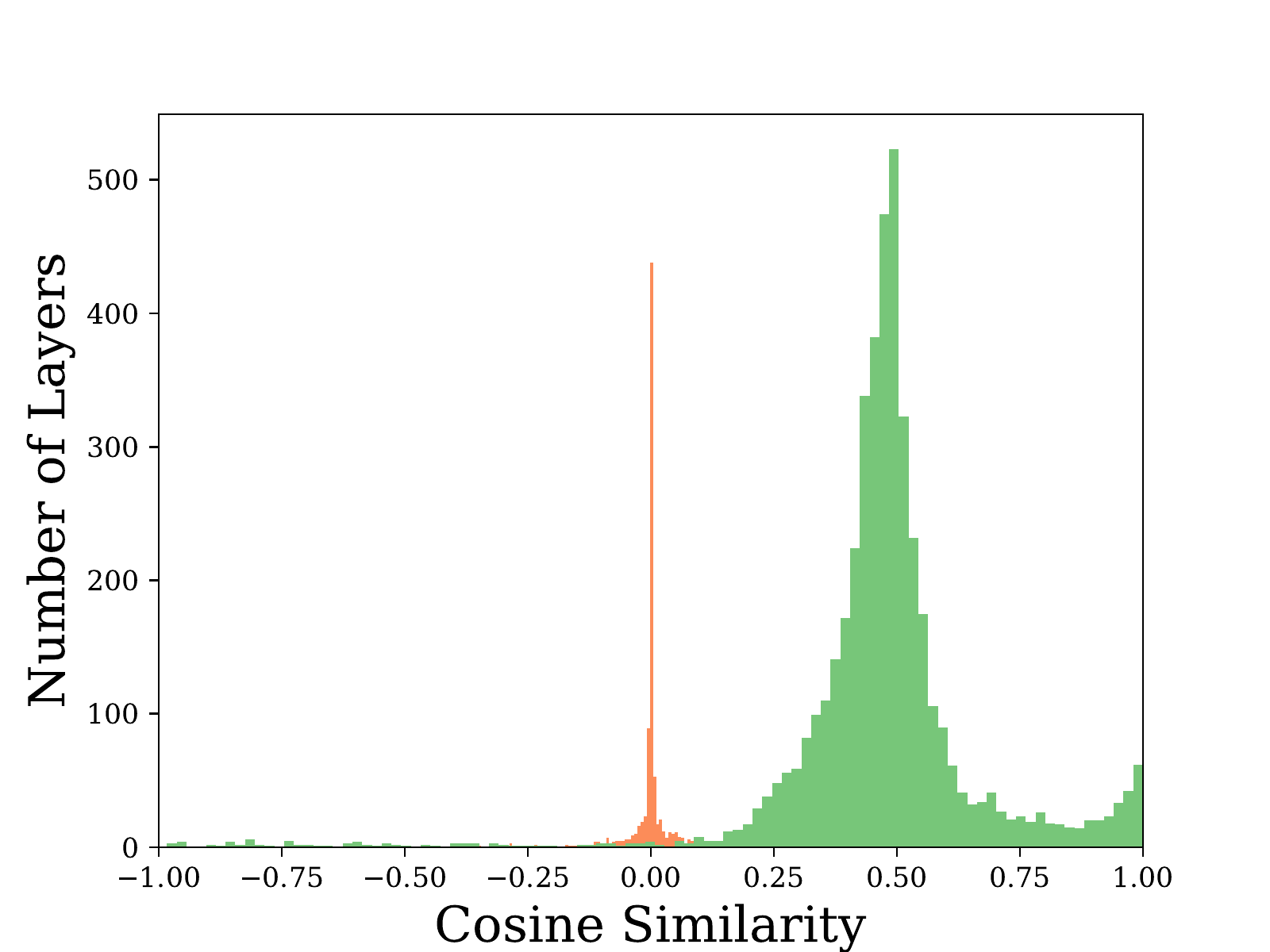}
    \includegraphics[width=0.49\textwidth]{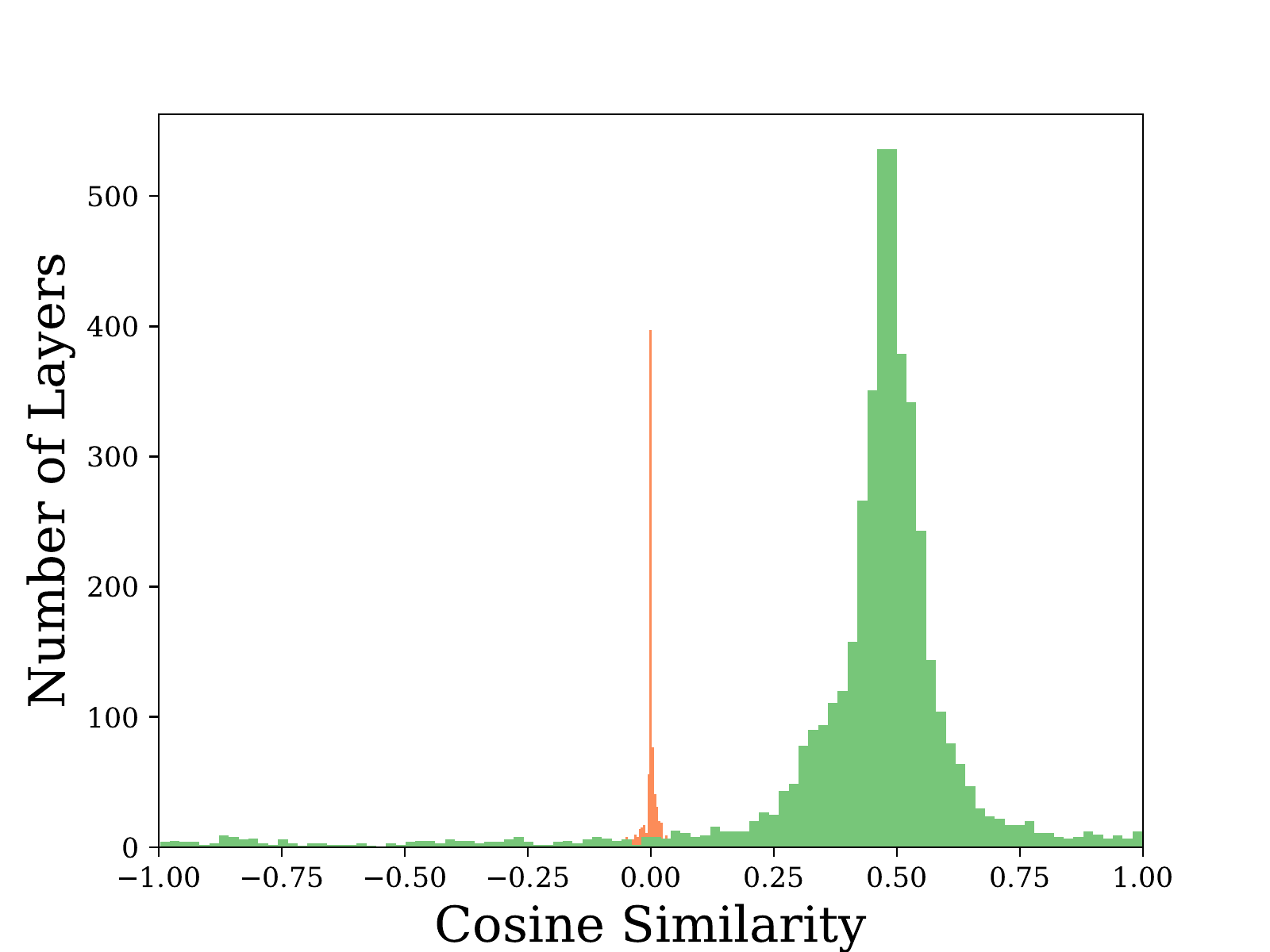}
    \caption{For the two examples shown, we compare the cosine similarity between the $\Delta\theta$ vectors learned by \methodshort{} (colored in orange above) and the vectors learned by decade transformations $\theta_t - \theta_i$ (colored in green above). We notice that the cosine similarity is centered around zero. This implies that in StyleGAN's parameter space, the \methodshort{} offset has no correlation with the decade transformation direction. To contrast, when comparing two decade offsets $\theta_{t_1} - \theta_i$ and $\theta_{t_2} - \theta_i$ we see that the vectors have high similarity. This agrees with our qualitative results, where we observe that \methodshort{} offset improves the identity preservation in each decade independently, without sacrificing the style of the target decade.}
    \label{fig:pti_appendix}
\end{figure*}
\end{document}